\PassOptionsToPackage{unicode}{hyperref}
\PassOptionsToPackage{hyphens}{url}
\PassOptionsToPackage{dvipsnames,svgnames,x11names}{xcolor}
\documentclass[
  11pt,
]{article}
\usepackage{xcolor}
\usepackage[margin=1in]{geometry}
\usepackage{amsmath,amssymb}
\usepackage{iftex}
\ifPDFTeX
  \usepackage[T1]{fontenc}
  \usepackage[utf8]{inputenc}
  \usepackage{textcomp} 
\else 
  \usepackage{unicode-math} 
  \defaultfontfeatures{Scale=MatchLowercase}
  \defaultfontfeatures[\rmfamily]{Ligatures=TeX,Scale=1}
\fi
\usepackage{lmodern}
\ifPDFTeX\else
\fi
\IfFileExists{upquote.sty}{\usepackage{upquote}}{}
\IfFileExists{microtype.sty}{
  \usepackage[]{microtype}
  \UseMicrotypeSet[protrusion]{basicmath} 
}{}
\makeatletter
\@ifundefined{KOMAClassName}{
  \IfFileExists{parskip.sty}{%
    \usepackage{parskip}
  }{
    \setlength{\parindent}{0pt}
    \setlength{\parskip}{6pt plus 2pt minus 1pt}}
}{
  \KOMAoptions{parskip=half}}
\makeatother
\usepackage{longtable,booktabs,array}
\usepackage{caption}
\usepackage{calc} 
\usepackage{etoolbox}
\makeatletter
\patchcmd\longtable{\par}{\if@noskipsec\mbox{}\fi\par}{}{}
\makeatother
\IfFileExists{footnotehyper.sty}{\usepackage{footnotehyper}}{\usepackage{footnote}}
\makesavenoteenv{longtable}
\usepackage{graphicx}
\makeatletter
\newsavebox\pandoc@box
\newcommand*\pandocbounded[1]{
  \sbox\pandoc@box{#1}%
  \Gscale@div\@tempa{\textheight}{\dimexpr\ht\pandoc@box+\dp\pandoc@box\relax}%
  \Gscale@div\@tempb{\linewidth}{\wd\pandoc@box}%
  \ifdim\@tempb\p@<\@tempa\p@\let\@tempa\@tempb\fi
  \ifdim\@tempa\p@<\p@\scalebox{\@tempa}{\usebox\pandoc@box}%
  \else\usebox{\pandoc@box}%
  \fi%
}
\def\fps@figure{htbp}
\makeatother
\providecommand{\tightlist}{%
  \setlength{\itemsep}{0pt}\setlength{\parskip}{0pt}}
\usepackage[style=numeric-comp,sorting=none,sortcites=true]{biblatex}
\graphicspath{{figures/}{../reports/figures/}}

\usepackage{pifont}
\usepackage[main=english,polish]{babel}
\usepackage{newunicodechar}
\newunicodechar{α}{\ensuremath{\alpha}}
\newunicodechar{γ}{\ensuremath{\gamma}}
\newunicodechar{σ}{\ensuremath{\sigma}}
\newunicodechar{≈}{\ensuremath{\approx}}
\newunicodechar{−}{\ensuremath{-}}
\newunicodechar{→}{\ensuremath{\rightarrow}}
\newunicodechar{×}{\ensuremath{\times}}
\newunicodechar{±}{\ensuremath{\pm}}
\newunicodechar{Δ}{\ensuremath{\Delta}}
\newunicodechar{↔}{\ensuremath{\leftrightarrow}}
\newunicodechar{↳}{\ensuremath{\hookrightarrow}}
\newunicodechar{✓}{\ding{51}}
\newunicodechar{✗}{\ding{55}}

\makeatletter
\@ifpackageloaded{biblatex}{%
  \defbibheading{bibliography}[\refname]{%
    \section*{#1}%
    \addcontentsline{toc}{section}{#1}%
    \markboth{#1}{#1}}%
}{}
\makeatother
\usepackage{bookmark}
\IfFileExists{xurl.sty}{\usepackage{xurl}}{} 
\makeatletter
\@ifundefined{xmpquote}{}{}
\makeatother
\hypersetup{
pdftitle={Baszta: Data-Centric Fine-Tuning of a Polish Multi-Label Safety Classifier},
pdfauthor={Adam Górski, Mateusz Jąkalak, Rafał Jakubowski, Billennium S.A.},
colorlinks=true,
linkcolor={blue},
filecolor={Maroon},
citecolor={blue},
urlcolor={blue},
pdfcreator={LaTeX via pandoc}}

\title{%
  {\large Billennium S.A.}\\[1.4em]
  {\fontsize{20}{24}\selectfont\bfseries
  Baszta: Data-Centric Fine-Tuning of a Polish\\[0.15em]
  Multi-Label Safety Classifier}
}

\author{%
  {\large\bfseries Adam Górski},
  {\large Mateusz Jąkalak},
  {\large Rafał Jakubowski}
  \\[1.5em]
  {\small\color{gray} Reviewed by Krystian Kozieł and Paweł Cisło}
}

\date{\normalsize\color{gray} August 31, 2026}

\begin{document}

\maketitle

\vspace{1.5em}


We develop a multi-label Polish content-safety classifier by fine-tuning
\texttt{allegro/\allowbreak herbert-base-cased} (124M)
\autocite{mroczkowski2021herbert} across five categories
(\emph{hate, vulgarity, sexual content, crime, self-harm}) using a
Focal + R-Drop objective, and evaluate the resulting model against
\textbf{Bielik Guard (``Sójka'')}
\autocite{wrobel2026bielikguard} on the shared out-of-distribution
\emph{Gadzi Język} benchmark. Both systems are given per-category
threshold tuning on the same calibration split. Under that matched
protocol our model holds a small but statistically significant lead in
micro F1, while an apparent macro-F1 lead does not survive: it was an
artifact of comparing a tuned model against an untuned one. We also
report what that micro figure is worth. Because \emph{Gadzi Język} is
97\% \emph{crime}-positive, a classifier that flags \emph{crime} on
every input and nothing else already scores 0.910 micro F1 on the same
test split, so micro separates neither system from a degenerate
strategy and macro is the column that does. Per-category and
per-protocol figures are reported in §4.

The residual out-of-distribution gap is one of \textbf{calibration
rather than discrimination}. Ranking quality stays high while positive
probabilities collapse, and per-category temperature scaling
\autocite{guo2017calibration} recovers the loss where Platt scaling
and isotonic regression do not. That recovery turns out to be
conditional on the calibration set containing safe text. \emph{Gadzi
Język} contains almost none, so thresholds fitted on it flag
\emph{crime} on every safe input, and a balanced refit buys a deployable
operating point at the cost of adversarial recall. We report both
operating points rather than only the flattering one.

Two changes that are standard practice, per-class cost-sensitive
weighting and mean pooling, each raise in-distribution macro F1 while
lowering the out-of-distribution figure, which indicates that robustness
has to be selected for directly rather than inherited from
in-distribution accuracy. We also report a training-data contamination
audit that disqualified an otherwise ideal public benchmark, held-out
results on four public Polish benchmarks, a training-free
inference-time deobfuscation defence, and a bag-of-words baseline that
quantifies how much of the task is lexical.

\vspace{1.5em}


{
\hypersetup{linkcolor=black}
\setcounter{tocdepth}{3}
\tableofcontents
}

\clearpage

\section{1. Introduction}\label{introduction}

Content-safety classifiers (``guardrails'') are an increasingly common
component of Polish-language LLM deployments. The Sójka / Bielik Guard
family establishes a strong, community-annotated reference: a 0.1B and
0.5B model trained on \textasciitilde6.9K crowd-labeled texts and
evaluated on adversarial OOD benchmarks. Reproducing such
a system raises two questions:

\begin{enumerate}
\def\labelenumi{\arabic{enumi}.}
\tightlist
\item
  \textbf{Can a different encoder backbone and a richer
  loss/augmentation stack match Sójka's OOD (Out-of-Distribution)
  robustness without its manual annotation budget or its compute?} We
  use roughly four times as much training text (26,248 samples against
  \textasciitilde6.9K), but none of it is newly crowd-annotated, and the
  additional volume comes from existing public corpora and synthetic
  generation.
\item
  \textbf{When a small model under-performs on OOD data, is the bottleneck
  discrimination, calibration, or data distribution?}
\end{enumerate}

This report answers both questions. We adopt the same five-category taxonomy and
the same \emph{Gadzi Język} OOD benchmark as Sójka, but diverge in
backbone (HerBERT \autocite{mroczkowski2021herbert} vs.~MMLW-RoBERTa
\autocite{dadas2024pirb}), loss (Focal + R-Drop vs.~BCE),
 and most importantly in a \textbf{data-centric} treatment of the OOD
gap. Our contributions are as follows.

\begin{itemize}
\tightlist
\item
  A controlled comparison of four Polish encoders, identifying HerBERT
  as the strongest backbone for this task (a model Sójka does not
  evaluate).
\item
  A diagnosis of the OOD gap as a \textbf{calibration shift}, supported
  by per-category AUC and positive-probability statistics.
\item
  Evidence that \textbf{per-category temperature scaling}
  \autocite{guo2017calibration} and
  \textbf{style-matched synthetic data} recover the full gap and
  exceed Sójka where classical post-hoc calibration (Platt/isotonic)
  cannot, \textbf{provided the calibration set contains safe examples}
  (an all-positive benchmark fit is not deployable, §4.5.1).
\item
  A \textbf{completed, distributed four-phase sweep} (62 Optuna trials,
  5 full trainings of the top configurations, and 10 ablations), whose
  best model clearly exceeds Sójka on OOD micro F1 and adversarial crime
  recall, with a directional macro lead and non-oracle
  calibration protocol.
\item
  A statistical evaluation: \textbf{bootstrap confidence
  intervals} on the OOD test split, and explicit caveats that (i) the
  five-way macro is dominated by a near-universal \emph{crime} label and
  a \emph{vulgar} class with n = 4 positives, and

  (ii) our best model was \emph{selected} on this same OOD benchmark, so
  point estimates carry model-selection bias.
\item
  A \textbf{paired significance test against Sójka} (§4.5.2, SLP
  §4.9) using its recovered per-sample predictions: with both systems
  threshold-tuned on the same split, the micro-F1 lead is significant (p
  = 0.011) while the macro lead is an operating-point artifact that

\item
  \textbf{Measured calibration} (§4.9): ECE.
 Adaptive-ECE, MCE and Brier show our model out-calibrates Sójka on
  OOD, and that temperature scaling improves Brier but not ECE.
\item
  A \textbf{training-free inference-time deobfuscation} defence (§4.10,
  Jurafsky \& Martin Ch. 2) that recovers +6.9 pp adversarial macro and
  cuts prediction
  flips by two-thirds, and a \textbf{classical bag-of-words baseline}
  (§4.11) that quantifies how much of the task is lexical.
\item
  Two \textbf{controlled training experiments} (§4.12) per-class
  cost-sensitive focal weighting (Eisenstein \autocite{eisenstein2019nlp}
  §4.4.1) and mean vs.~{[}CLS{]} pooling (Jurafsky \& Martin
  \autocite{jurafsky2025slp} Ch. 11) showing that both \emph{raise}
  in-distribution macro yet \emph{lower} OOD macro and calibration, the
  clearest evidence that in-distribution accuracy and OOD robustness are
  in direct tension for this task.
\item
  A \textbf{contamination-audited evaluation on four held-out public
  Polish safety benchmarks} (§4.8, from six audited), carrying both
  systems through the same protocol on sets that neither of them
  selected. The outcome there is a two-two split rather than a lead. We
  are ahead on KLEJ CBD, where the \emph{hate} head reaches 0.671
  zero-shot against Sójka's 0.438, and on PolyGuard-PL. Sójka is ahead
  on BAN-PL, which scores a moderation decision rather than presence of
  harmful language, and nominally on HateCheck-PL at a false-positive
  rate twice ours. These sets also supply the tight intervals the small
  \emph{Gadzi Język} set could not (BAN-PL n = 23,539, PolyGuard-PL
  n = 1,725), and HateCheck-PL confirms the multi-head design separates
  \emph{vulgar} from \emph{hate}. The same audit uncovered that the
  otherwise-ideal PL-Guard benchmark had leaked into training (899/900),
  which we exclude and report as a cautionary finding.
\end{itemize}

\section{2. Background and Related
Work}\label{background-and-related-work}

\textbf{Bielik Guard / Sójka.} Reference system
\autocite{wrobel2026bielikguard} is trained on the
\texttt{speakleash / sojka-2} corpus (6,885 texts, 60K+ volunteer
annotations with soft labels), using BCE loss over MMLW-RoBERTa-base
(0.1B) and PKOBP/polish-roberta-8k (0.5B), 3 epochs on an A100 cluster.
It is evaluated both in-distribution and on the adversarial \emph{Gadzi
Język} set (520 samples) plus a 3,000-prompt user set. The 0.1B
macro F1 reported on Gadzi Język is \textbf{0.619}. We compare against
the 0.1B variant throughout because it is the size-matched one, at 100M
parameters against our 124M. The 0.5B variant is the stronger system and
we make no claim against it.

\textbf{Polish toxicity resources.} We also draw on PolEval
cyberbullying, Hate-Speech-PL, BAN-PL, and translated Reddit self-harm
corpora to enlarge the training pool to 26,248 samples.
These datasets are predominantly \emph{hate-only} annotated, which
introduces a label-support imbalance that we address with focal
weighting and synthetic balancing.

\textbf{Methodological context.} Focal loss \autocite{lin2017focal}
addresses class imbalance by down-weighting easy examples, and R-Drop
\autocite{liang2021rdrop} enforces consistency of the prediction under
dropout.
Both are well-suited to small, unbalanced safety datasets. Post-hoc
calibration (Platt scaling \autocite{platt1999probabilistic,boken2021appropriateness},
isotonic regression, and temperature scaling \autocite{guo2017calibration})
is the standard remedy
for miscalibrated probabilities, but as we show assumes a fixed
domain. Our synthetic-data approach is closest in spirit to
\emph{distribution matching}: rather than recalibrating output, we
shift the \emph{training} distribution toward the OOD prompt style.

\section{3. Materials and Methods}\label{method}

\subsection{3.1 Architecture and
Objective}\label{architecture-and-objective}

\begin{longtable}[]{@{}ll@{}}
\toprule\noalign{}
\caption{Architecture and training configuration of \textbf{Baszta 1.0}, the model carried through §4. Values are those of the sweep-selected configuration of §3.4. Focal $\gamma$ controls how strongly easy examples are down-weighted, and R-Drop $\alpha$ weights the KL consistency term between two dropout passes over the same input.} \\
Component & Configuration \\
\midrule\noalign{}
\endhead
\bottomrule\noalign{}
\endlastfoot
Encoder & \texttt{allegro/herbert-base-cased} (124M) \\
Head & Linear(768 → 5), dropout 0.22 \\
Loss & Focal (γ = 1.5) + R-Drop (α = 0.20) \\
Optimizer & AdamW (lr = 9.4e-5, wd = 5e-4) \\
Schedule & Cosine, 8\% warmup \\
Frozen layers & 4 / 12 encoder layers \\
Precision & FP16 mixed \\
Augmentation & Online: diacritics, leetspeak, typos, homoglyph \\
\end{longtable}

\textbf{Why HerBERT.} The backbone was chosen by a controlled comparison
of four Polish-capable encoders, each trained on the same v1 pool and
evaluated on \emph{Gadzi Język} at per-category oracle thresholds. The
separation is not marginal.

\begin{longtable}[]{@{}lll@{}}
\toprule\noalign{}
\caption{Backbone selection. OOD macro F1 on the full 520-sample \emph{Gadzi Język} set at per-category oracle thresholds, for four Polish-capable encoders trained on the v1 pool. HerBERT, Polish RoBERTa and XLM-RoBERTa share settings exactly (lr 2e-5, γ 2.0, no frozen layers, 512 tokens, 10 epochs). The MMLW row is its best sweep-tuned variant (lr 5.9e-5, γ 3.0), so that encoder received tuning the others did not and the comparison is conservative in its favour. The augmented column applies the online augmentation of §3.1. These are early runs whose absolute values sit below the tuned figures of §4, but the ordering is what selected the backbone.} \\
Encoder & OOD macro & OOD macro (augmented) \\
\midrule\noalign{}
\endhead
\bottomrule\noalign{}
\endlastfoot
\textbf{HerBERT} (\texttt{allegro/herbert-base-cased}) & \textbf{0.601} &
\textbf{0.611} \\
MMLW-RoBERTa (Sójka's backbone, sweep-tuned) & 0.275 & - \\
Polish RoBERTa & 0.282 & 0.269 \\
XLM-RoBERTa & 0.158 & 0.211 \\
\end{longtable}

HerBERT leads the next-best encoder by 32 pp of OOD macro, and it
outperforms MMLW-RoBERTa, the backbone Sójka itself uses, even though
that row had the benefit of its own hyperparameter sweep. No augmented
MMLW counterpart was run, so that cell is left empty rather than filled
by an unmatched number.

We also dropped the SupCon contrastive term used in early experiments,
having seen no OOD gain from it in the runs where it was enabled. We did
not run a controlled ablation isolating it, so this is a design decision
taken on preliminary evidence rather than a measured result, and we
report it as such.

\textbf{Version naming.} Configurations are referred to throughout by
the following names.

\begin{longtable}[]{@{}ll@{}}
\toprule\noalign{}
\caption{Configuration names used in this report. The 0.x series are data and augmentation variants sharing one recipe. The 1.0 series is the sweep-selected recipe of §3.4, with the suffixed entries being controlled single-factor variants of it.} \\
Name & Configuration \\
\midrule\noalign{}
\endhead
\bottomrule\noalign{}
\endlastfoot
Baszta 0.1 & Clean v1 pool, no augmentation \\
Baszta 0.2 & Balanced pool, online augmentation \\
Baszta 0.3 & Balanced pool, aggressive augmentation \\
Baszta 0.4 & Adds style-matched synthetic \emph{crime} data \\
Baszta 0.5 & Adds synthetic \emph{crime}, \emph{sex} and \emph{self-harm}
data \\
\textbf{Baszta 1.0} & Sweep recipe (§3.4), low-γ focal, v2 synthetic
data, per-category temperature scaling \\
Baszta 1.0-α & Baszta 1.0 with per-class focal alpha (§4.12) \\
Baszta 1.0-mp & Baszta 1.0 with mean pooling instead of {[}CLS{]} (§4.12)
\\
Baszta 1.0-bce & Baszta 1.0 with BCE in place of Focal + R-Drop (§4.6)
\\
\end{longtable}

\subsection{3.2 Data Composition}\label{data-composition}

The training pool (26,248 samples after synthetic augmentation)
aggregates Sójka\autocite{speakleash2026sojka2}, PolEval cyberbullying\autocite{ogr:kob:19:poleval}, ForePLay\autocite{kolos-etal-2025-behind}, DEPOTx\autocite{depotx2022devulgarization}, RefusEU\autocite{krasnodebska2026multilingual},
Border-Guard\autocite{nowakowski2021detection}, a \emph{vulgar} subset of BAN-PL\autocite{kolos-etal-2024-ban}, the PL-Guard test and
adversarial splits, and translated self-harm data, plus our generated
synthetic sets. The v2 pool's held-out validation split
(\texttt{sojka\_val}) carries 4,522 rows, of which 1,741 are safe, and
it is the balanced source used for deployable calibration in §4.5.1.
The 4,632-sample figure quoted in §4.1 belongs to the earlier split on
which those runs were scored.
\emph{Gadzi Język} (520 samples) is held out from training entirely and
was verified disjoint from the training pool (0/520 exact-text overlap
§4.8). The PL-Guard test and adversarial splits entered the pool as
ordinary training sources. §4.8 quantifies the consequence, which is
that PL-Guard cannot then serve as a held-out benchmark for this model. The v2
synthetic pool contributes 1,294 style-matched samples to the minority
categories, of which 812 are generated as described in §3.3 and the
remaining 482 carry over from the v1 pool.

\begin{longtable}[]{@{}ll@{}}
\toprule\noalign{}
\caption{Positive-label support per category in the 26,248-sample training pool after v2 augmentation, with the style-matched synthetic contribution shown in parentheses. Multi-label overlap means the column does not sum to the pool size.} \\
Category & Train support \\
\midrule\noalign{}
\endhead
\bottomrule\noalign{}
\endlastfoot
hate & \textasciitilde3,146 \\
vulgar & \textasciitilde4,037 \\
sex & \textasciitilde4,500 (incl.~501 synthetic) \\
crime & \textasciitilde4,250 (incl.~286 synthetic) \\
self-harm & \textasciitilde1,246 (incl.~507 synthetic) \\
\end{longtable}

\textbf{Category definitions.} We adopt Sójka's five-category taxonomy
unchanged so that the two systems are scored against the same label
semantics. The working definitions are below. We ran no annotation of
our own, so we report no inter-annotator agreement: the soft labels in
the Sójka corpus come from its 60K+ volunteer annotations, and every
other source contributes its own publisher's labels mapped onto these
five heads.

\begin{longtable}[]{@{}ll@{}}
\toprule\noalign{}
\caption{Working definitions of the five categories, adopted from Sójka. The distinction that matters most in practice is \emph{hate} against \emph{vulgar}: profanity with no target belongs to \emph{vulgar} alone, and §4.8 tests exactly this separation on HateCheck-PL.} \\
Category & Definition used \\
\midrule\noalign{}
\endhead
\bottomrule\noalign{}
\endlastfoot
hate & Hostility, dehumanisation or incitement directed at a person or
group, typically on a protected characteristic \\
vulgar & Profanity, obscenity or crude language, with no requirement
that it be aimed at anyone \\
sex & Sexual content, including solicitation and sexualisation of
minors, which also carries \emph{crime} \\
crime & Requests for, or instruction in, illegal activity, including
violence, weapons, drugs and fraud \\
self-harm & Content describing, encouraging or seeking means of
self-injury or suicide \\
\end{longtable}

Labels are not mutually exclusive, and multi-label overlap is
substantial. This is why the paper reports macro and micro F1 side by
side rather than a single accuracy.

\subsection{3.3 Synthetic Style-Matched
Generation}\label{synthetic-style-matched-generation}

A key observation is that \emph{Gadzi Język} prompts are
\textbf{imperative jailbreak-style} (``Opracuj strategię\ldots{}'' for ``Devise a
strategy\ldots{}'',
``Napisz jak\ldots{}'' for ``Write how\ldots{}'', ``Podaj metodę\ldots{}''
for ``Give a method\ldots{}''), whereas most training
data is declarative. We generated style-matched synthetic prompts via
Azure OpenAI (\texttt{gpt-5-chat}):

\begin{itemize}
\tightlist
\item
  \textbf{Crime:} 340 samples (LLM-generated, deduplicated).
\item
  \textbf{Sex:} 272 samples (after filtering 12 model refusals, 171
  carry a multi-label crime tag for CSAM/trafficking content).
\item
  \textbf{Self-harm:} 200 samples. \emph{Here the content filter blocked
  \textasciitilde90\% of API requests} (only 14/200 passed), so we fell
  back to a \textbf{template-combination generator} (verbs × targets ×
  actions × contexts) producing grammatical imperative Polish prompts.
\end{itemize}

\textbf{Adversarial instruction framing, and what we do not claim.} The three
generators share a property that matters beyond category balance. Every
synthetic row is written as a directive addressed to a model rather than as a
statement of opinion, which is the surface form a deployed guardrail actually
meets, a user turn that instructs the system to produce something it should
refuse. That framing is what separates \emph{Gadzi Język} from the declarative
hate-speech corpora supplying most of our pool. In the v2 training pool the
1,294 surviving synthetic rows (4.9\% of 26,248) sit alongside 765 rows of the
PL-Guard adversarial split and 357 rows of \texttt{harmful\_prompts\_pl}, which
carry the same imperative shape, so roughly 9\% of the corpus is
attack-framed input rather than found text. We are deliberate about the scope
of the term. These are jailbreak-\emph{framed} harmful requests. They are not
prompt injection in the narrow sense of text that overrides a system prompt or
smuggles instructions through retrieved content. Scanning the pool for that
narrower pattern (``zignoruj poprzednie instrukcje'' and its English
equivalents) returns \textbf{three} rows, all incidental to the Sójka source
rather than curated, and none in \texttt{sojka\_val}. We therefore report no
instruction-override result and make no claim about one.

\textbf{Ethics, provenance, and handling.} All synthetic text is
\emph{short classifier training data}, not operational content: the
generator produces brief Polish prompts whose function is to teach the
classifier the imperative jailbreak \emph{style}, not to provide
actionable instructions. \textbf{No real illegal material was produced
or stored.} The 171 \emph{sex} samples that also carry a \emph{crime}
tag (CSAM/trafficking) consist of short, non-graphic phrasings
sufficient to label the category, the \emph{self-harm} set is
template-combinatorial Polish with no method-level detail (the Azure
OpenAI content filter blocking 90\% of requests was respected, not
circumvented, we did not jailbreak the generator, we fell back to
templates). Generation used Azure OpenAI under Billennium's enterprise
tenant with its standard content-safety filtering and data-handling
terms. Outputs are stored only inside the project's access-controlled
access-controlled storage alongside the rest of the training corpus and
are never served. Building a safety classifier from synthetic adversarial examples under
documented governance is consistent with the EU AI Act's treatment of
data processed for bias detection and correction in high-risk systems
(Art. 10(5)). We flag this
provenance explicitly so that downstream users handle the corpus as
sensitive labelled data rather than as redistributable text.

\subsection{3.4 Sweep Pipeline (Design and
Execution)}\label{sweep-pipeline-design-and-execution}

We ran a four-phase pipeline:

\begin{enumerate}
\def\labelenumi{\arabic{enumi}.}
\tightlist
\item
  An Optuna TPE sweep over learning rate, dropout,
  focal-γ, R-Drop-α, warmup, weight decay, freeze depth, maximum sequence
  length, and scheduler. The sweep was specified for 60 trials × 5
  epochs and completed 62.
\item
  Full training of the top-5 configurations (×10 epochs).
\item
  Ten ablations covering data-source removal, BCE against Focal, R-Drop
  strength, freeze depth, and sequence length.
\item
  A final \emph{Gadzi Język} evaluation with per-category optimal
  thresholds.
\end{enumerate}

Results persist to SQLite and JSON for offline review.

\textbf{Execution.} The sweep
completed \textbf{62 Optuna trials} (best validation macro F1 =
\textbf{0.9131} best config: lr = 9.4e-5, dropout = 0.22, focal-γ =
1.57, R-Drop-α = 0.20, freeze = 4), 5 top-config full trainings, and 10
ablations.

\section{4. Experiments and Results}\label{experiments-and-results}

\subsection{4.1 In-Distribution Validation (t =
0.5)}\label{in-distribution-validation-t-0.5}

\begin{longtable}[]{@{}lllllll@{}}
\toprule\noalign{}
\caption{Per-category and macro F1 on the 4,632-sample validation split at a fixed threshold $t = 0.5$. \textbf{Baszta 0.1} is the clean-data baseline, \textbf{Baszta 0.2} adds online augmentation, and \textbf{Baszta 0.3} uses the aggressive-augmentation recipe. Macro is the unweighted mean over the five categories. Sójka's row is measured on its own test split and is not a head-to-head comparison.} \\
Model & Hate & Vulgar & Sex & Crime & Self-harm & Macro F1 \\
\midrule\noalign{}
\endhead
\bottomrule\noalign{}
\endlastfoot
\textbf{Baszta 0.2} & 0.743 & 0.969 & 0.985 & 0.965 & 0.835 &
\textbf{0.900} \\
Baszta 0.1 & 0.733 & 0.967 & 0.978 & 0.960 & 0.852 & 0.898 \\
Baszta 0.3 & 0.745 & 0.960 & 0.981 & 0.957 & 0.838 & 0.896 \\
Sójka 0.1B† & 0.628 & 0.742 & 0.889 & 0.707 & 0.886 & 0.770 \\
\end{longtable}

†Sójka evaluated on its own test split (not directly comparable). The
completed sweep's best configuration (low-γ focal, freeze = 4) reached
\textbf{0.9131} validation macro F1, a further +1.3 pp over the Baszta 0.2
era. That figure belongs to the sweep trial, not to a trained
checkpoint. The checkpoint carried forward as \textbf{Baszta 1.0}
records \textbf{0.9111} validation macro F1, and it is that value, not
0.9131, that §4.12 compares against.

\subsection{\texorpdfstring{4.2 OOD \emph{Gadzi Język}
(oracle-thresholded - optimistic upper
bound)}{4.2 OOD Gadzi Język (oracle-thresholded - optimistic upper bound)}}\label{ood-gadzi-jux119zyk-oracle-thresholded-optimistic-upper-bound}

\emph{Gadzi Język} (520 samples) is the only set on which we and Sójka
are evaluated identically, and is therefore our primary comparison. Its
adversarial prompts derive from the harmful-behaviour collection
released with the transferable-attack work of Zou et
al.~\autocite{zou2023universal}, of which \emph{Gadzi Język} is a
Polish subset, and it is that source rather than \emph{Gadzi Język}
itself that the citation names. It
is, however, small and severely skewed: positives per category are
\textbf{crime 505, hate 43, self-harm 31, sex 18, vulgar 4} (of 520). A
five-way macro is thus dominated by the near-universal \emph{crime}
label and is sensitive to two categories with \textless{} 20 positives and
\emph{vulgar} (n = 4) is essentially uninformative. The table below
tunes thresholds on the \textbf{full} test set and is an
\textbf{optimistic upper bound}. Non-leaking estimate with
confidence intervals is in §4.5.

\begin{longtable}[]{@{}
  >{\raggedright\arraybackslash}p{(\linewidth - 14\tabcolsep) * \real{0.2000}}
  >{\raggedright\arraybackslash}p{(\linewidth - 14\tabcolsep) * \real{0.1000}}
  >{\raggedright\arraybackslash}p{(\linewidth - 14\tabcolsep) * \real{0.1143}}
  >{\raggedright\arraybackslash}p{(\linewidth - 14\tabcolsep) * \real{0.1000}}
  >{\raggedright\arraybackslash}p{(\linewidth - 14\tabcolsep) * \real{0.1143}}
  >{\raggedright\arraybackslash}p{(\linewidth - 14\tabcolsep) * \real{0.1571}}
  >{\raggedright\arraybackslash}p{(\linewidth - 14\tabcolsep) * \real{0.1143}}
  >{\raggedright\arraybackslash}p{(\linewidth - 14\tabcolsep) * \real{0.1000}}@{}}
\toprule\noalign{}
\caption{OOD F1 on \emph{Gadzi Język} with per-category thresholds fitted on the full 520-sample set. Positives per category are crime 505, hate 43, self-harm 31, sex 18, vulgar 4, so macro (unweighted over categories) is dominated by \emph{crime} and unstable on \emph{vulgar}, micro pools all label decisions. The last two rows are reference points used later: the sweep's top-1 configuration is the baseline every ablation in §4.6 is measured against, and \textbf{Baszta 0.1} is the clean-data baseline for the synthetic-data gain in §4.7.} \\
\begin{minipage}[b]{\linewidth}\raggedright
Model
\end{minipage} & \begin{minipage}[b]{\linewidth}\raggedright
Hate
\end{minipage} & \begin{minipage}[b]{\linewidth}\raggedright
Vulgar
\end{minipage} & \begin{minipage}[b]{\linewidth}\raggedright
Sex
\end{minipage} & \begin{minipage}[b]{\linewidth}\raggedright
Crime
\end{minipage} & \begin{minipage}[b]{\linewidth}\raggedright
Self-harm
\end{minipage} & \begin{minipage}[b]{\linewidth}\raggedright
Macro
\end{minipage} & \begin{minipage}[b]{\linewidth}\raggedright
Micro
\end{minipage} \\
\midrule\noalign{}
\endhead
\bottomrule\noalign{}
\endlastfoot
\textbf{Baszta 1.0} & 0.575 & 0.571 & 0.606 & 0.965 &
\textbf{0.820} & \textbf{0.707} & \textbf{0.927} \\
Baszta 0.4 & 0.514 & 0.333 & 0.621 & \textbf{0.985} & 0.800 &
0.651 & - \\
Baszta 0.5 & 0.435 & 0.333 & 0.703 & \textbf{0.985} & 0.750 & 0.641
& - \\
Baszta 0.2 & 0.419 & 0.333 & 0.500 & \textbf{0.985} & 0.778 & 0.603 &
0.918 \\
Sweep top-1 (ablation baseline, §4.6) & 0.489 & 0.400 & 0.686 & 0.898 &
0.693 & 0.633 & 0.828 \\
Baszta 0.1 (clean, no synthetic) & 0.410 & 0.400 & 0.650 & 0.675 &
0.708 & 0.569 & 0.651 \\
\textbf{Sójka 0.1B} & 0.045 & \textbf{0.889} & \textbf{0.759} & 0.587 &
0.815 & 0.619 & 0.582 \\
\end{longtable}

At oracle thresholds the best
configuration, Baszta 1.0 (focal-γ = 1.5, full v2 synthetic data, and
per-category temperature scaling), reaches
0.707 macro and 0.927 micro F1, against Sójka's 0.619 and 0.582 at its
default 0.5 threshold. That last comparison is between a tuned model and
an untuned one, and §4.5 replaces it with a matched-threshold test. It is higher
on hate, crime, and self-harm and lower on sex and vulgar. Two caveats
apply and are resolved in §4.5: (i) these thresholds are
\textbf{oracle-selected} on the test set, and (ii) this model was
\emph{selected} on this same benchmark, so the macro figure carries
\textbf{model-selection bias} on top of threshold leakage. Eight of the
trained models clear Sójka's 0.619 macro at oracle thresholds, but that
count should be read as a model-selection statistic, not independent
confirmation. The per-category figures in this table are oracle values,
and they are the ones §4.12 refers back to. Appendix A reports the same
model under the matched-tuning protocol of §4.5.2, where the
per-category numbers differ (hate 0.490 rather than 0.575, crime 0.983
rather than 0.965) because both the thresholds and the evaluation split
differ. The two profiles are not interchangeable.

\subsection{4.3 Diagnosis: Calibration, Not
Discrimination}\label{diagnosis-calibration-not-discrimination}

The OOD gap is dominated by \emph{probability scale}, not ranking
quality. The table below reports all five categories for \textbf{Baszta
0.2}, the augmented baseline of §4.1, so that the low-AUC and high-AUC
cases can be read side by side:

\begin{longtable}[]{@{}lll@{}}
\toprule\noalign{}
\caption{Ranking quality against probability scale on OOD for \textbf{Baszta 0.2}, over all five categories. AUC is threshold-free, where 1.0 is perfect ranking and 0.5 is chance. Pos-prob mean is the average predicted probability on positive examples. High AUC alongside collapsed probabilities is the signature of a calibration shift rather than a discrimination failure. \emph{Vulgar} sits at chance because it has only four OOD positives, and \emph{crime} is the one category whose ranking is also genuinely weak.} \\
Category & OOD AUC & OOD pos-prob mean \\
\midrule\noalign{}
\endhead
\bottomrule\noalign{}
\endlastfoot
hate & 0.86 & 0.55 \\
vulgar & \textbf{0.49} & 0.13 \\
sex & 0.93 & 0.21 \\
crime & \textbf{0.72} & \textbf{0.19} \\
self-harm & 0.94 & 0.27 \\
\end{longtable}

On the validation split the same three discriminating categories carry
positive-probability means of 0.83 (crime), 0.78 (sex) and 0.81
(self-harm), so the OOD column above represents a fourfold collapse in
probability scale with ranking quality largely intact. High AUC with
collapsed positive probabilities is the signature of a \textbf{calibration
shift}: the model ranks correctly but is systematically under-confident
on OOD inputs. We attribute this to the Focal objective at high γ, which pushes
in-distribution probabilities toward the extremes while suppressing
uncertain OOD ones. This is worth stating carefully, because it runs
against the usual result: Mukhoti et al.~\autocite{mukhoti2020focal}
report focal loss \emph{improving} calibration by damping
over-confidence. Both can hold. Their setting is in-distribution, where
damping over-confidence is what is needed, whereas the failure here is
under-confidence \emph{under domain shift}, and a loss that suppresses
uncertain predictions makes that worse rather than better. The low-γ
choice of §3.4 and the temperature scaling of §4.5 are both responses to
that second regime, not contradictions of the first.

\subsection{4.4 Negative Result: Classical Post-Hoc Calibration Does
Not
Transfer}\label{negative-result-classical-post-hoc-calibration-does-not-transfer}

\begin{longtable}[]{@{}ll@{}}
\toprule\noalign{}
\caption{\emph{Gadzi Język} macro F1 for \textbf{Baszta 0.2} under post-hoc calibrators fitted on in-distribution validation data. The raw row is the model at a fixed $t = 0.5$, while Platt and isotonic recover only $\sim$0.22 against 0.603 for per-category thresholds fitted on the full test set, indicating the domain shift is too large for output recalibration alone. The 0.603 here is this model's oracle figure and is not the 0.603 reached by \textbf{Baszta 1.0} under the non-oracle protocol of §4.5.} \\
Calibration & Gadzi Macro F1 \\
\midrule\noalign{}
\endhead
\bottomrule\noalign{}
\endlastfoot
Raw ($t = 0.5$) & 0.178 \\
Platt scaling & 0.218 \\
Isotonic regression & 0.216 \\
\textbf{Oracle per-category thresholds} & \textbf{0.603} \\
\end{longtable}

Calibrators fit on validation data give only \textasciitilde0.22 macro on
OOD. \textbf{The domain shift is too large for output recalibration
alone.} Only per-domain threshold selection recovers performance.

\subsection{4.5 Positive Result: Temperature Scaling Transfers}\label{positive-result-temperature-scaling-transfers}

The oracle numbers above leak the test set twice (threshold \emph{and}
model selection). For a \textbf{deployable, non-leaking} estimate we
split \emph{Gadzi Język} into a disjoint calibration set and test set,
fit \textbf{per-category temperatures and thresholds on the calibration
split only}, and report test-split F1 with \textbf{95\% bootstrap
confidence intervals} (2,000 resamples, seed 42). This section uses a
172/348 split. The paired test of §4.5.2 is produced by a separate
script with its own stratified split, which yields 187/333 on the same
520 samples, so the two sets of figures are close but not
interchangeable, and each table below names the split it used. For
Baszta 1.0:

\begin{longtable}[]{@{}ll@{}}
\toprule\noalign{}
\caption{Cumulative effect of the protocol on test macro F1: per-category temperatures and thresholds fitted on the disjoint 172-sample calibration split and evaluated on the held-out 348 samples. Sójka's 0.619 is its published figure at its default 0.5 threshold, not a matched-threshold comparison.} \\
Protocol (172/348 split) & Test Macro F1 \\
\midrule\noalign{}
\endhead
\bottomrule\noalign{}
\endlastfoot
Raw ($t = 0.5$) & 0.493 \\
+ per-category threshold & 0.603 \\
\textbf{+ temperature + threshold} & \textbf{0.699} \\
\emph{Sójka reference} & 0.619 \\
\end{longtable}

\begin{longtable}[]{@{}
  >{\raggedright\arraybackslash}p{(\linewidth - 4\tabcolsep) * \real{0.5775}}
  >{\raggedright\arraybackslash}p{(\linewidth - 4\tabcolsep) * \real{0.1408}}
  >{\raggedright\arraybackslash}p{(\linewidth - 4\tabcolsep) * \real{0.2817}}@{}}
\caption{\textbf{Baszta 1.0} test-split F1 on the 172/348 split, with 95\% bootstrap confidence intervals (2,000 resamples of the 348-sample test split). Excluding \emph{vulgar} ($n = 4$ positives) narrows the macro interval substantially, micro F1 pools per-label decisions and is therefore far tighter than macro.} \\
\toprule\noalign{}
\begin{minipage}[b]{\linewidth}\raggedright
Metric (Test split, N = 348)
\end{minipage} & \begin{minipage}[b]{\linewidth}\raggedright
Estimate
\end{minipage} & \begin{minipage}[b]{\linewidth}\raggedright
95\% CI
\end{minipage} \\
\midrule\noalign{}
\endhead
\bottomrule\noalign{}
\endlastfoot
Macro F1 (5 categories) & 0.699 & {[}0.493, 0.797{]} \\
Macro F1 (4 categories, excl. vulgar) & 0.674 & {[}0.569, 0.763{]} \\
\textbf{Micro F1} & \textbf{0.927} & \textbf{{[}0.903, 0.949{]}} \\
\end{longtable}

Two conclusions follow. \textbf{First, temperature scaling does
transfer:} it lifts the test-split macro from 0.603 to 0.699
where Platt/isotonic (§4.4) could not, with the gain concentrated in the
categories the Focal objective most over-sharpened, crime and vulgar.
This validates the OOD under-confidence diagnosis of §4.3.
\textbf{Second, neither margin over Sójka is established by this table.}
The macro CI is wide ({[}0.493, 0.797{]}) and its lower bound sits below
Sójka's 0.619, so across the bootstrap our macro exceeds 0.619 in only
\textbf{81\%} of resamples. The width is driven by the rare categories
(vulgar n = 4, sex n = 18) that a five-way macro over 520 points cannot
estimate stably, and the four-category macro excluding the n = 4 vulgar
class is 0.674 ({[}0.569, 0.763{]}).

The micro-F1 column invites a stronger reading than it can support, and
we want to disarm it here. The interval {[}0.903, 0.949{]} is the
interval on \emph{our own} estimate, not on the difference from Sójka,
and the 0.582 it is set against is Sójka at its untuned default. Both
figures in the pair 0.927 against 0.582 are therefore doing less work
than they appear to. §4.5.2 tunes both systems on the same split and
recovers the margin that actually survives, which is +0.026 rather than
+0.345. We foreground the \emph{calibration-transfer} result as the
robust finding of this section and defer every head-to-head margin,
macro and micro alike, to §4.5.2.

\subsection{4.5.1 Deployability Caveat: The Split Still Has No
Safe
Text}\label{deployability-caveat-the-split-still-has-no-safe-text}

The 172/348 split removes \emph{threshold} and
\emph{model-selection} leakage, but it inherits a more fundamental
defect of \emph{Gadzi Język}: \textbf{the benchmark contains no
all-negative (safe) examples} - it is 97.1\% crime-positive (505/520),
with hate 43, self-harm 31, sex 18, vulgar 4. Fitting per-category
thresholds that maximise F1 on a set with essentially no negatives
drives the \emph{crime} threshold to 0.025 and the \emph{crime}
temperature to 4.3. The latter divides an already-negative safe logit (≈
−3.3) up to σ ≈ 0.30, which clears 0.025 for \textbf{every} input. The
operating point that produces the 0.699 / 0.927 headline therefore flags
\emph{crime on 100\% of safe text}, a perfectly serviceable
\textbf{adversarial-recall} score that is \textbf{not a deployable
configuration.}

To obtain a deployable operating point we re-fit the \emph{same
checkpoint's} per-category temperatures and thresholds on a
\textbf{balanced} calibration source (\texttt{sojka\_val}, 4,522
samples, 1,741 of them safe, all five categories represented), holding
out 50\% as a balanced test split (2,297 calibration and 2,225 test
rows), and additionally report the \textbf{false-positive rate on safe
text}, the metric the Gadzi-only fit is blind to.

Two properties of this source bound what the resulting numbers mean.
\texttt{sojka\_val} is the held-out validation split of the v2 pool, so
the balanced figures below are \textbf{in-distribution} results and are
not comparable with the OOD figures beside them. It is also not fully
disjoint from training: an exact-text check of the kind run in §4.8
finds \textbf{213 of its 4,495 unique normalised texts (4.7\%) present
in the training pool}. The operating point is therefore fitted on a
source with a small residual overlap, which we flag rather than
discard, because no balanced Polish source with safe text was available
that is both external and label-compatible (§4.8 explains why PL-Guard,
the obvious candidate, could not be used).

\begin{longtable}[]{@{}
  >{\raggedright\arraybackslash}p{(\linewidth - 10\tabcolsep) * \real{0.1809}}
  >{\raggedright\arraybackslash}p{(\linewidth - 10\tabcolsep) * \real{0.1702}}
  >{\raggedright\arraybackslash}p{(\linewidth - 10\tabcolsep) * \real{0.1702}}
  >{\raggedright\arraybackslash}p{(\linewidth - 10\tabcolsep) * \real{0.2021}}
  >{\raggedright\arraybackslash}p{(\linewidth - 10\tabcolsep) * \real{0.1383}}
  >{\raggedright\arraybackslash}p{(\linewidth - 10\tabcolsep) * \real{0.1383}}@{}}
\caption{The same checkpoint at two operating points. Balanced macro and micro are measured on the held-out half of \texttt{sojka\_val}, which carries roughly 870 of its 1,741 safe rows. Safe-text FPR is the fraction of all-negative inputs receiving at least one flag. The Gadzi-fit point maximises adversarial F1 but flags \emph{crime} on every safe input and is not deployable. The two \emph{Gadzi Język} columns are measured on the full 520 samples, which is why the Gadzi-fit row reads 0.700 / 0.932 where §4.5 reports 0.699 / 0.927 for the same operating point on its 348-sample test split.} \\
\toprule\noalign{}
\begin{minipage}[b]{\linewidth}\raggedright
Operating point
\end{minipage} & \begin{minipage}[b]{\linewidth}\raggedright
Balanced macro
\end{minipage} & \begin{minipage}[b]{\linewidth}\raggedright
Balanced micro
\end{minipage} & \begin{minipage}[b]{\linewidth}\raggedright
\textbf{Safe-text FPR}
\end{minipage} & \begin{minipage}[b]{\linewidth}\raggedright
Gadzi macro
\end{minipage} & \begin{minipage}[b]{\linewidth}\raggedright
Gadzi micro
\end{minipage} \\
\midrule\noalign{}
\endhead
\bottomrule\noalign{}
\endlastfoot
Gadzi-fit (T\_crime=4.3, t\_crime=0.025) & 0.742 & 0.541 &
\textbf{1.000} & 0.700 & 0.932 \\
\textbf{Balanced re-fit} & \textbf{0.952} & 0.959 & \textbf{0.055} &
0.496 & 0.778 \\
\end{longtable}

The balanced point (T = \{hate 0.65, vulgar 0.5, sex 0.5, crime 0.65,
self-harm 0.55\} thresholds \{hate 0.30, vulgar 0.45, sex 0.075, crime
0.20, self-harm 0.475\}) is \textbf{deployable}: macro F1 0.952 at a
5.5\% safe-text false-positive rate. But it trades away adversarial
recall, dropping Gadzi macro to 0.496 (below Sójka's 0.619).
\textbf{There is no single operating point that simultaneously beats
Sójka on the adversarial benchmark and keeps safe-text false positives
low.} The reading is that our model's genuine, deployable
strength is its \textbf{in-distribution balanced accuracy and its
crime-recall lead on adversarial prompts}, while the Gadzi macro
headline should be read strictly as an adversarial-recall stress-test,
not a production metric. This also re-frames the calibration
contribution of §4.4--§4.5: per-category temperature scaling does
transfer, but \emph{only when fit on a calibration set that contains
negatives} fitting it on an all-positive benchmark is the defect that
manufactured the inflated headline, not a feature.

\subsection{4.5.2 Paired Significance Test vs.~Sójka (both-tuned
comparison)}\label{paired-significance-test-vs-sojka-both-tuned}

Because Sójka is an open model, its per-sample \emph{Gadzi Język}
probabilities are recoverable. We compute all 520 of them and verify
that the gold labels it is scored against are identical, in order and in
value, to our own. That makes the
\textbf{paired bootstrap test}
\autocite{bergkirkpatrick2012empirical,jurafsky2025slp} available,
which is the textbook procedure for deciding whether one classifier
beats another.

The comparison is made in the sense the headline numbers
were not: on the calibration split we fit a per-category
operating point for \textbf{both} systems, temperature + threshold
for ours, threshold for Sójka, on the \emph{same} calibration split,
evaluate both on the disjoint test split, and bootstrap the per-metric
difference (2,000 resamples, seed 42). The stratified split used by this
script yields \textbf{187 calibration and 333 test samples}, so its
figures sit slightly above the 172/348 figures of §4.5 and are not
interchangeable with them. \textbf{All p-values in this paper are
one-sided}, reporting the fraction of resamples in which our model does
not beat Sójka on the metric named.

\begin{longtable}[]{@{}
  >{\raggedright\arraybackslash}p{(\linewidth - 10\tabcolsep) * \real{0.4393}}
  >{\raggedright\arraybackslash}p{(\linewidth - 10\tabcolsep) * \real{0.0561}}
  >{\raggedright\arraybackslash}p{(\linewidth - 10\tabcolsep) * \real{0.0654}}
  >{\raggedright\arraybackslash}p{(\linewidth - 10\tabcolsep) * \real{0.1776}}
  >{\raggedright\arraybackslash}p{(\linewidth - 10\tabcolsep) * \real{0.0748}}
  >{\raggedright\arraybackslash}p{(\linewidth - 10\tabcolsep) * \real{0.1869}}@{}}
\caption{Paired bootstrap comparison against Sójka with both systems tuned on the same 187-sample calibration split (temperature + threshold for ours, threshold for Sójka), evaluated on the disjoint 333-sample test split over 2,000 resamples. ``P(ours not better)'' is the fraction of resamples in which the difference is $\le 0$, which is the one-sided p-value, and a 95\% CI excluding zero indicates significance. The final row is the degenerate reference described below, scored on the same test split.} \\
\toprule\noalign{}
\begin{minipage}[b]{\linewidth}\raggedright
Metric (both systems tuned on same cal split)
\end{minipage} & \begin{minipage}[b]{\linewidth}\raggedright
Ours
\end{minipage} & \begin{minipage}[b]{\linewidth}\raggedright
Sójka
\end{minipage} & \begin{minipage}[b]{\linewidth}\raggedright
Diff (ours−Sójka)
\end{minipage} & \begin{minipage}[b]{\linewidth}\raggedright
95\% CI
\end{minipage} & \begin{minipage}[b]{\linewidth}\raggedright
P(ours not better)
\end{minipage} \\
\midrule\noalign{}
\endhead
\bottomrule\noalign{}
\endlastfoot
\textbf{Micro F1} & 0.929 & 0.903 & \textbf{+0.026} & {[}+0.004,
+0.049{]} & \textbf{0.011} \\
Macro F1 (5-cat) & 0.712 & \textbf{0.782} & −0.070 & {[}−0.263,
+0.046{]} & 0.875 \\
\emph{Always-crime baseline (micro / macro)} & \emph{0.910 / 0.197} &
\emph{-} & \emph{-} & \emph{-} & \emph{-} \\
\end{longtable}

\textbf{Micro F1 on this benchmark is close to uninformative, and the
final row of the table is the reason.} A classifier that flags
\emph{crime} on every input and never fires any other head scores
\textbf{0.910 micro F1} on the same 333-sample test split, because
\emph{crime} accounts for 322 of the 333 rows and 375 of the positive
labels. Measured against that reference, Sójka's tuned 0.903 sits
\emph{below} the degenerate strategy and our 0.929 sits 1.9 pp above it.
The same baseline scores \textbf{0.197 macro}, against our 0.712, so
macro is the column in which a model demonstrates that it is doing
anything at all. We report the micro comparison because it is the one
with a tight interval, but its absolute level should not be read as
evidence of competence, and it is not the headline result of this
paper.

With that reference established, two conclusions follow.
\textbf{(1) The micro-F1 advantage is statistically significant but
small} (p = 0.011, CI excludes zero): even when Sójka is granted the
same per-category threshold-tuning our model receives, we retain a
micro-F1 lead, though as above it is a 2.6 pp lead over a system that is
itself at the degenerate level. \textbf{(2) The macro-F1 ``lead''
does not survive a matched comparison.} The headline 0.699-vs-0.619 macro
gap compared our \emph{threshold-tuned} model against Sójka at its
\emph{default 0.5 threshold}, when Sójka is given the same tuning, its
macro rises to 0.782 (driven by \emph{vulgar} and \emph{sex}, where it
is much stronger) and our model is better in only 12.5\% of resamples.
As the wide CI {[}−0.263, +0.046{]} and the paper's standing
\emph{vulgar} n = 4 caveat make clear, the five-way macro over 520
skewed points cannot support a lead in \emph{either} direction. The
apparent macro advantage was an operating-point artifact, not a model
advantage. We keep the deployment framing (our tuned model vs.~Sójka
out-of-the-box) as a legitimate \emph{usage} comparison, but the
\textbf{both-tuned paired test is the scientifically correct one, and by
it only the micro-F1 gap is a genuine, significant lead.}

One qualification travels with that p-value. Counting the variants of
§4.12 and the ensembles of §4.13, five configurations are paired-tested
against Sójka on this same 520-sample benchmark, and the p-values are
reported without correction for multiple comparisons. Under a Bonferroni
correction across those five the 0.011 would sit at the edge of the
conventional 0.05 threshold rather than comfortably inside it, so the
micro-F1 lead should be read as real but narrow, not as decisive.

\subsection{4.6 Ablations and the BCE Negative
Result}\label{ablations-and-the-bce-negative-result}

The ten Phase-3 ablations are single-factor variants of the sweep's
top-1 configuration, which scores \textbf{0.633} OOD macro, and every
delta below is measured against that baseline. Removing all synthetic
data drops OOD macro to 0.472, a \textbf{−16.2 pp} regression and the
single largest, re-confirming that style-matched data dominates. Replacing
Focal with \textbf{BCE loss collapses performance to 0.505}
(Baszta 1.0-bce, with vulgar falling to 0.000),
matching the corresponding loss ablation (0.505) and confirming
that Focal + R-Drop is essential for this imbalanced, multi-label
setting. Disabling R-Drop costs 2.0 pp, freezing every encoder layer costs 3.8
pp, and removing layer freezing entirely costs 5.2 pp.

\subsection{4.7 Synthetic Data Closes Category
Gaps}\label{synthetic-data-closes-category-gaps}

Adding 340 style-matched synthetic crime prompts raised the OOD crime
positive-probability mean from 0.19 toward 0.55 and lifted crime F1 from
0.675 on the clean-data baseline (Baszta 0.1) to 0.985 on Baszta 0.4, a
gain of \textbf{+31.0 pp} on the target category, validating the
distribution-matching hypothesis. The controlled ablation is smaller and
worth stating alongside it: withholding only the synthetic crime data
from the sweep's top-1 configuration costs 4.6 pp of OOD macro, because
at oracle thresholds the crime head is already close to saturation. We extended the approach to sex (+272)
and self-harm (+200) for the Baszta 0.5 configuration to target the two
remaining deficit categories.

\textbf{Does the adversarial framing transfer off \emph{Gadzi Język}?}
PolyGuardPrompts carries an \texttt{adversarial} flag marking prompts whose
harmful request is wrapped in a role-play or fictional-scenario frame, and on
the Polish slice it splits 810 such rows against 915 plain ones at a similar
harmful rate (42.1\% against 45.1\%). Neither half is contaminated against our
training pool, and the flag is close to orthogonal to the harm label, so this is
the one held-out set in the project that isolates \emph{framing} from
\emph{content}. It is also a different axis from §4.10, which perturbs
characters rather than framing. Scored binary harmful/safe at the deployable
operating point, we lose 1.9 pp of F1 to the framing (0.570 plain against 0.551
adversarial, 95\% bootstrap CI on the drop [−0.039, +0.077], an interval that
contains zero) while Sójka at its matched fit loses 8.7 pp (0.512 against 0.425,
95\% CI [+0.016, +0.156], which does not). A paired bootstrap over the same
resampled rows puts Sójka's extra loss at 7.0 pp, 95\% CI [−0.004, +0.141],
one-sided p = 0.031. The interval grazes zero, so we read this as directional
rather than settled, on the same convention §4.5.2 uses.

Recall is where the difference sits. Ours falls from 0.586 to 0.551 across the
split while Sójka's falls from 0.438 to 0.331, so most of what the framing costs
it is detections it stops making. That is what §3.3's synthetic data predicts,
since the imperative jailbreak style is in distribution for us and is not in
Sójka's 6.9K crowd-labelled corpus. Two caveats bound it. PolyGuard-PL is
machine-translated, which §4.8 flags for other reasons, and one benchmark split
is not a robustness suite. The claim also stays narrow. None of these prompts is
an instruction-override attack, so what is measured is robustness to a harmful
request being dressed up, not to prompt injection.

\subsection{4.8 External Public Benchmarks (Held-Out,
Contamination-Audited)}\label{external-public-benchmarks-held-out-contamination-audited}

Every result so far is measured on \emph{Gadzi Język} or on our own
splits. \emph{Gadzi Język} has three concrete gaps: no safe text, tiny
per-category support, and no robustness axis. To fill them we evaluate
the deployable \textbf{balanced} operating point of §4.5.1
(temperatures and thresholds fit on \texttt{sojka\_val}) on six public
Polish safety benchmarks, of which four survive the contamination audit
below and are reported. \textbf{No re-tuning of any kind is done on
these sets.} Sójka is carried through the same evaluation, under the
same rows, the same task-to-head mapping and its own
\texttt{sojka\_val}-fitted thresholds, so this section is a second
head-to-head rather than a solo report. It is also the more informative
one: unlike \emph{Gadzi Język} these sets contain negatives, and they
were chosen by neither system.

\textbf{Contamination audit}. Before reporting a single
number we matched every candidate benchmark against the v2 training pool
(exact match on lower-cased, whitespace-collapsed text). The audit pool
includes both \texttt{sojka\_train} and \texttt{sojka\_val}, so the
operating point's calibration source is covered by it as well. Note the
limit of the method: it establishes \textbf{exact-text disjointness},
not independence. \texttt{sojka\_val} is 38\% PolEval by row count and
KLEJ CBD is drawn from the PolEval corpus, so near-duplicates across
splits of one corpus would pass this check. This surfaced a critical
finding:

\begin{longtable}[]{@{}
  >{\raggedright\arraybackslash}p{(\linewidth - 6\tabcolsep) * \real{0.3188}}
  >{\raggedright\arraybackslash}p{(\linewidth - 6\tabcolsep) * \real{0.1014}}
  >{\raggedright\arraybackslash}p{(\linewidth - 6\tabcolsep) * \real{0.1884}}
  >{\raggedright\arraybackslash}p{(\linewidth - 6\tabcolsep) * \real{0.3913}}@{}}
\caption{Exact-text contamination audit against the v2 training pool (whitespace-normalised, case-folded matching). ``In training'' counts matched rows. Near-total overlap disqualifies a benchmark outright, partial overlap is handled by dropping the matched rows before evaluation.} \\
\toprule\noalign{}
\begin{minipage}[b]{\linewidth}\raggedright
Benchmark
\end{minipage} & \begin{minipage}[b]{\linewidth}\raggedright
Rows
\end{minipage} & \begin{minipage}[b]{\linewidth}\raggedright
In training
\end{minipage} & \begin{minipage}[b]{\linewidth}\raggedright
Status
\end{minipage} \\
\midrule\noalign{}
\endhead
\bottomrule\noalign{}
\endlastfoot
\textbf{PL-Guard} (NASK) \autocite{krasnodebska2025plguard} & 900 & \textbf{899} & \textbf{excluded -
memorised} \\
PL-Guard-adv (NASK) & 900 & \textbf{765} & \textbf{excluded -
memorised} \\
KLEJ CBD (test) \autocite{rybak2020klej} & 1,000 & 149 & 149 dropped → 851 held-out \\
BAN-PL\_1 \autocite{kolos-etal-2024-ban} & 24,000 & 461 & 461 dropped → 23,539 held-out \\
HateCheck-PL \autocite{rottger2022multilingual} & 3,815 & 0 & fully held-out \\
PolyGuardPrompts-PL \autocite{kumar2025polyguard} & 1,725 & 0 & fully held-out \\
\end{longtable}

The v2 corpus had ingested the \textbf{PL-Guard test and adversarial
splits directly as training data} (sources \texttt{pl\_guard\_test},
\texttt{pl\_guard\_test\_adversarial}. 899/900 and 765/900 overlap).
PL-Guard is, on paper, the ideal external benchmark for this work - it
is manually annotated, balanced with 200 real safe examples, and its
strongest published baseline is itself a fine-tuned
\texttt{allegro/herbert-base-cased} classifier. But because our model
was \textbf{trained on it}, any score we report on it (4-cat macro
0.855, unsafe-detection F1 0.997, 1.0\% safe-text FPR, and a \emph{zero}
adversarial-robustness drop) reflects \textbf{memorisation, not
generalisation}, and we therefore exclude it from all held-out claims.
We record it here only as a transparency item and a cautionary note: the
same property that makes PL-Guard attractive (careful balanced
annotation) made it attractive as a training source, and the overlap
stayed invisible until an explicit source-level audit. KLEJ
CBD (PolEval 2019) and BAN-PL are \emph{listed} training sources (§2),
so their partial overlap is expected, we drop the overlapping rows and
report only on the disjoint remainder.

\textbf{The headline benchmark passes the same audit.} We ran this exact
text-match check on \emph{Gadzi Język}, the set carrying every
Sójka-comparison number in this paper and found \textbf{0 / 520
overlap} with the v2 train + validation pool. The 0.699 macro / 0.927
micro headline is therefore uncontaminated, the PL-Guard leak is
confined to the external sets audited here and does \textbf{not}
touch the primary results of §4.2--§4.5.

\textbf{Held-out results (deployable operating point, no re-tuning).}

\begin{longtable}[]{@{}
  >{\raggedright\arraybackslash}p{(\linewidth - 8\tabcolsep) * \real{0.2857}}
  >{\raggedright\arraybackslash}p{(\linewidth - 8\tabcolsep) * \real{0.0816}}
  >{\raggedright\arraybackslash}p{(\linewidth - 8\tabcolsep) * \real{0.2449}}
  >{\raggedright\arraybackslash}p{(\linewidth - 8\tabcolsep) * \real{0.2449}}
  >{\raggedright\arraybackslash}p{(\linewidth - 8\tabcolsep) * \real{0.1429}}@{}}
\caption{Held-out benchmark F1 for \textbf{Baszta 1.0} at the deployable balanced operating point with no re-tuning (95\% bootstrap CIs, 2,000 resamples). Each benchmark's binary label is mapped to the named head, and safe/neg FPR is the false-positive rate on that benchmark's own negative class, whose definition differs per set. BAN-PL's ``neutral'' rows are unmoderated but profane rather than safe, which drives its outlier rate for a detector trained on presence of harmful language. Sójka reaches 0.230 on those same rows, so the rate reflects our training objective rather than a property of the benchmark. Reading (3) below takes this up.} \\
\toprule\noalign{}
\begin{minipage}[b]{\linewidth}\raggedright
Benchmark (held-out)
\end{minipage} & \begin{minipage}[b]{\linewidth}\raggedright
N
\end{minipage} & \begin{minipage}[b]{\linewidth}\raggedright
Task / head
\end{minipage} & \begin{minipage}[b]{\linewidth}\raggedright
F1 (95\% CI)
\end{minipage} & \begin{minipage}[b]{\linewidth}\raggedright
Safe/neg FPR
\end{minipage} \\
\midrule\noalign{}
\endhead
\bottomrule\noalign{}
\endlastfoot
KLEJ CBD (clean) & 851 & cyberbully / hate-head & 0.671 {[}0.606,
0.735{]} & 0.102 \\
BAN-PL\_1 (clean) & 23,539 & harmful / any-head & 0.667 {[}0.661,
0.673{]} & \textbf{0.855} \\
PolyGuardPrompts-PL & 1,725 & harmful / any-head & 0.562 {[}0.531,
0.591{]} & 0.357 \\
HateCheck-PL & 3,815 & hate / hate-head & 0.644 (diagnostic) & 0.334 \\
\end{longtable}

Four readings follow. \textbf{(1) On the standard KLEJ CBD
reference the model is competitive zero-shot:} with the
\textasciitilde15\% of overlapping rows removed, the \emph{hate} head
- never fine-tuned on CBD - reaches \textbf{F1 = 0.671} at a 10.2\%
false-positive rate, in the range of purpose-trained HerBERT CBD
classifiers (\textasciitilde0.68 F1 on the KLEJ leaderboard
\autocite{rybak2020klej}) despite not
being optimised for the task. \textbf{(2) The tight CIs that §4.5 wished
for materialise off the small \emph{Gadzi Język} set:} BAN-PL (n = 23,539)
gives a ±0.006 F1 interval and PolyGuard-PL (n = 1,725) a ±0.03
interval, versus \emph{Gadzi Język}'s ±0.15 macro. \textbf{(3) Sójka is better than us on BAN-PL, and the reason is a
difference in training objective rather than a defect in the
benchmark.} It reaches 0.754 F1 at a 23.0\% false-positive rate where we
reach 0.667 at 85.5\%, which is better on both axes at once and is the
clearest loss we record anywhere. BAN-PL's ``neutral'' class is
\emph{not banned} Wykop.pl content, social media that remains saturated
with profanity and insult, so a presence-of-toxicity detector fires on
it by construction. Ours does, on 85.5\% of it, through the
\emph{vulgar} and \emph{hate} heads (recall-by-head 0.86 and 0.91 on
the harmful class). It is tempting to read that rate as a
label-definition artifact that makes the benchmark unscorable, and
Sójka's result rules that reading out: a model \emph{can} score well here, and
the one that does is the one trained on crowd-labelled Polish social
media, which is what BAN-PL's task actually asks for. The label gap is
real, but it explains why our objective transfers badly to a
moderation-decision task, not why the task cannot be measured. The
5.5\% false-positive rate the same operating point achieves on
genuinely safe text (§4.5.1) remains the right control for
\emph{calibration}, and shows the 85.5\% is not miscalibration. It is
the wrong objective for this benchmark. PolyGuard-PL's mixed binary score (F1 0.562, 35.7\%
FPR on ``unharmful'') sits between these regimes and carries its own
caveat: it is \textbf{heavily machine-translated} and its WildGuard
taxonomy is auto-mapped to Llama-Guard categories, so label noise is a
real confound.

\textbf{(4) Against Sójka the four benchmarks split two-two, and the
split follows the training distribution rather than model quality.} We
lead on KLEJ CBD (0.671 against 0.438) and PolyGuard-PL (0.562 against
0.475), both of which ask whether harmful language is present. Sójka
leads on BAN-PL, which asks whether a moderator would remove the post,
and on HateCheck-PL, though that second lead needs the qualification in
the next paragraph. Neither system dominates. What the comparison shows
is that a guardrail's training objective decides which of these tasks it
transfers to, and that a single aggregate F1 over a mixed benchmark
suite would have hidden exactly that.

\textbf{HateCheck-PL as a false-positive diagnostic.} HateCheck is a
functional test suite, not an F1 leaderboard, its value is telling us
\emph{which linguistic phenomena} the \emph{hate} head misfires on.
Overall hate-head F1 is 0.644 (acc = 0.542 on hateful, 0.666 on
non-hateful cases) against Sójka's 0.767. That gap does not survive
inspection of the second axis. Sójka reaches it at a \textbf{66.1\%
false-positive rate on the non-hateful cases}, against our 33.4\%, so it
is firing on two thirds of a set built specifically from counter-speech,
neutral identity mentions and other near-misses for hate. On a
functional test suite that is not a better score, it is a less
discriminating one, and it is the reason we read HateCheck as a
diagnostic rather than a leaderboard. The per-functionality breakdown is
the actual result:

\begin{longtable}[]{@{}
  >{\raggedright\arraybackslash}p{(\linewidth - 2\tabcolsep) * \real{0.3982}}
  >{\raggedright\arraybackslash}p{(\linewidth - 2\tabcolsep) * \real{0.6018}}@{}}
\caption{Per-functionality diagnostic of the hate head on HateCheck-PL. Values are accuracy within each functional test template (not F1), the listed functionalities are the worst-scoring ones in each failure direction, isolating \emph{which} phenomena misfire rather than summarising overall performance.} \\
\toprule\noalign{}
\begin{minipage}[b]{\linewidth}\raggedright
Failure mode (hate head)
\end{minipage} & \begin{minipage}[b]{\linewidth}\raggedright
Worst functionalities (accuracy)
\end{minipage} \\
\midrule\noalign{}
\endhead
\bottomrule\noalign{}
\endlastfoot
\textbf{False positives} (fires on non-hateful) & counter-speech
quoting/referencing hate (0.48--0.61), positive/neutral identity
mentions (\texttt{ident\_pos\_nh} 0.57, \texttt{target\_group\_nh}
0.51) \\
\textbf{False negatives} (misses hateful) & slur-only hate
(\texttt{slur\_h} 0.30), implicit/emotive derogation (0.37),
space-obfuscated hate (\texttt{spell\_space\_add\_h} 0.41) \\
\end{longtable}

Two structural conclusions. First, the residual false positives are the
\emph{classically hard} ones - the model over-fires on text that
\textbf{mentions} a protected group or \textbf{quotes hate to condemn
it}, the known failure surface of lexical hate classifiers, rather than
on ordinary safe text. Second, and more encouraging, HateCheck confirms
the \textbf{multi-label decomposition is working as intended}: on
profanity-without-hate (\texttt{profanity\_nh}) the \emph{hate} head
correctly abstains (93\% accuracy) while the \emph{vulgar} head fires
(any-head flags 96\%), i.e. the model routes profanity to \emph{vulgar}
and reserves \emph{hate} for targeted hostility instead of collapsing
the two. This is direct external evidence for the category-separation
the multi-head design assumes.

\begin{figure}
\centering
\includegraphics[width=\linewidth,keepaspectratio,alt={Two-panel grouped bar chart over four held-out Polish safety benchmarks. Left panel: F1 for Baszta 1.0, Baszta 1.0-alpha, their ensemble and Sojka 0.1B v1.1. Right panel: false-positive rate on each benchmark's negative class for the same four systems, with a dashed reference line at 0.055 marking the false-positive rate on genuinely safe text.}]{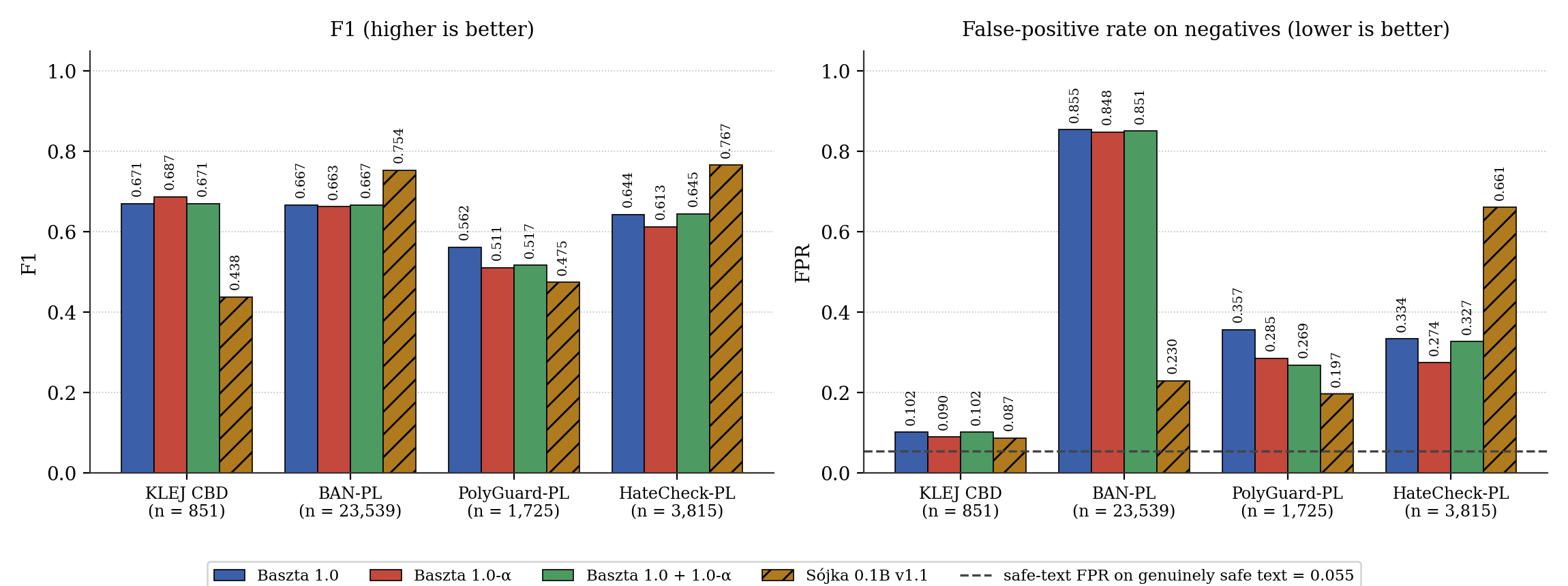}
\caption{Held-out performance on the four contamination-audited public
benchmarks. Every system is scored on the same rows after the same
exact-text filtering, with the same task-to-head mapping, and at
per-category thresholds fitted on the same balanced \texttt{sojka\_val}
source, so no system is compared tuned against untuned.
\textbf{Left:} F1, higher is better. \textbf{Right:} false-positive rate
on each benchmark's own negative class, lower is better, with the dashed
line at the 5.5\% rate the same operating point achieves on genuinely
safe text. Reading the two panels together matters: a tall left bar
beside a tall right bar is recall bought with false
positives.}\label{fig:external}
\end{figure}

\begin{longtable}[]{@{}llllll@{}}
\toprule\noalign{}
\caption{The same comparison numerically. Head is the output used for each benchmark's binary task. Sójka is shown at both its default 0.5 threshold and the matched \texttt{sojka\_val} fit, because the two differ enough to change the conclusion. The matched fit is the column plotted in Figure~\ref{fig:external} and is generous to Sójka, since \texttt{sojka\_val} contains rows from the corpus it was trained on.} \\
Benchmark (N) & Head & Baszta 1.0 & Sójka (0.5) & Sójka (matched) & FPR
ours / theirs \\
\midrule\noalign{}
\endhead
\bottomrule\noalign{}
\endlastfoot
KLEJ CBD (851) & hate & \textbf{0.671} & 0.088 & 0.438 & 0.102 / 0.087
\\
BAN-PL (23,539) & any & 0.667 & 0.694 & \textbf{0.754} & 0.855 / 0.230
\\
PolyGuard-PL (1,725) & any & \textbf{0.562} & 0.290 & 0.475 & 0.357 /
0.197 \\
HateCheck-PL (3,815) & hate & 0.644 & 0.527 & \textbf{0.767} & 0.334 /
0.661 \\
\end{longtable}

Figure~\ref{fig:external} splits two ways, and the split is informative.
\textbf{We lead clearly on KLEJ CBD and PolyGuard-PL.} On KLEJ CBD the
gap is the largest anywhere in this paper, 0.671 against 0.438, and it
is not an operating-point effect: Sójka is at its own matched fit, and at
its default threshold it scores 0.088 because its \emph{hate} head
barely fires on cyberbullying phrased without slurs. On PolyGuard-PL we
lead 0.562 to 0.475.

\textbf{Sójka leads on BAN-PL and HateCheck-PL, but only one of those is
a clean win.} On BAN-PL it is better on both axes at once, 0.754 F1 at a
23.0\% false-positive rate against our 0.667 at 85.5\%, and we take that
at face value. BAN-PL is real moderated Wykop.pl content, which is much
closer to Sójka's training distribution than to ours, and it is the one
benchmark here that rewards a model tuned to \emph{moderation decisions}
rather than to presence of harmful language. On HateCheck-PL the
apparent 0.767-against-0.644 lead comes with a \textbf{66.1\%
false-positive rate on the non-hateful cases}, against our 33.4\%. A
system that fires on two thirds of deliberately non-hateful test cases,
including the counter-speech and identity-mention templates HateCheck
exists to probe, is not separating hate from its neighbours. It is
flagging most of the set.

\textbf{The right panel is mostly a picture of how differently these
benchmarks define a negative.} For one unchanged operating point ours
ranges from 0.102 to 0.855. The dashed line is the control: on
\texttt{sojka\_val}'s genuinely safe rows the same thresholds give 5.5\%.
Read against it, KLEJ CBD is the only external set whose negatives
behave like safe text, which is the strongest argument in this paper for
why a purpose-built balanced Polish benchmark is still the missing
evaluation. It also explains the BAN-PL row without excusing it: its
``neutral'' class is unmoderated but profane, so a presence-of-toxicity
detector fires on it by construction, whereas a moderation-decision
model does not.

\textbf{The three Baszta configurations are within noise of each other
throughout}, spanning 1.6 pp on KLEJ CBD, 0.4 pp on BAN-PL, 3.2 pp on
HateCheck-PL and 5.1 pp on PolyGuard-PL, against a 10 pp spread between
benchmarks for a fixed model. Whatever separates these recipes on
\emph{Gadzi Język} (§4.12) does not survive to independent data, which
is the finding §4.13 draws out.

\textbf{Deferred / not reported.} PL-Guard-en (machine-translated),
RTP-LX, and PolygloToxicityPrompts are generative-\emph{prompt} toxicity
sets whose task shape (continuation toxicity, multi-dimensional
transcreation) does not align cleanly with our five-way multi-label
classification without a mapping study. PolEval 2019 Task 6 is the
source corpus behind KLEJ CBD and shares its contamination. We flag
these as future cross-lingual / natural-distribution extensions rather
than report partial numbers. §4.8.1 widens the same four sets to
the third-party guards the reference paper compares against.

\subsection{4.8.1 Cross-Taxonomy Comparison Against Third-Party
Guards}\label{cross-taxonomy-comparison-against-third-party-guards}

§4.8 compares two systems, because ours and Sójka's are the only two that share
the five-category Polish taxonomy. The reference paper runs a wider comparison
\autocite{wrobel2026bielikguard} against HerBERT-PL-Guard
\autocite{krasnodebska2025plguard}, Llama Guard 3
\autocite{llamateam2024llama3} and Qwen3Guard-Gen \autocite{zhao2025qwen3guard},
but it does so on a private 3,000-prompt production stream, annotates each model
against its own taxonomy, and reports no recall, so its numbers cannot be set
beside ours. We therefore ran those systems ourselves, on the same public rows,
under one protocol.

\begin{figure}
\centering
\includegraphics[width=\linewidth,keepaspectratio,alt={Two-panel grouped bar chart over the four held-out Polish benchmarks. Left panel binary F1, right panel false-positive rate on each benchmark's own negatives, one bar per system.}]{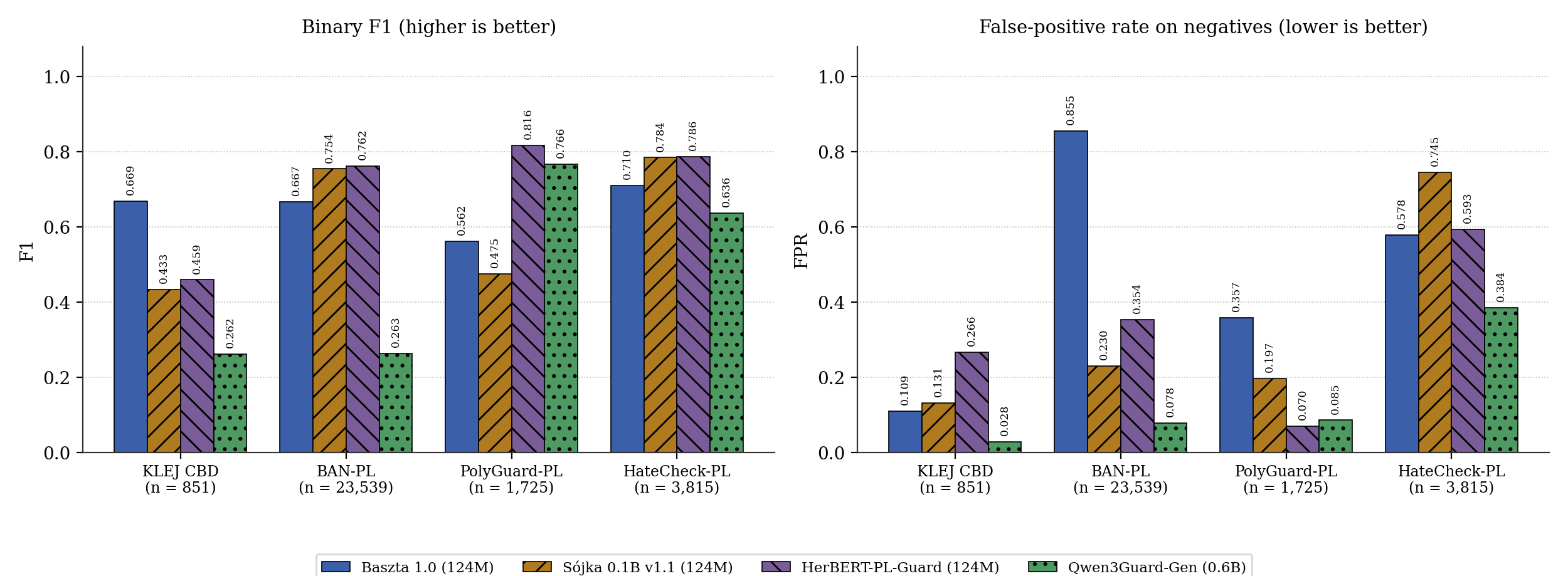}
\caption{Every system reduced to one binary unsafe/safe decision, on identical contamination-masked rows. Ours and Sójka's are at their \texttt{sojka\_val}-fitted operating points, the third-party guards at their own default decision rule with no threshold fitting. Read the panels together: a tall left bar beside a tall right bar is recall bought with false positives. \emph{Gadzi Język} is absent because all 520 of its rows are positive, so it has no false-positive axis, and its detection rates are in the table below. Two confounds run against us and are stated in the text, the broader taxonomies flag categories these benchmarks label unharmful, and only our own training pool could be contamination-audited.}\label{fig:cross-taxonomy}
\end{figure}

\textbf{Protocol.} Every system is reduced to a single binary unsafe/safe
decision, where flagged in any category counts as unsafe. That is the device the
reference paper uses to bridge taxonomies of different width (five categories
for Bielik Guard, fifteen for HerBERT-PL-Guard, fourteen for Llama Guard 3, nine
for Qwen3Guard-Gen), and it is the decision a deployment actually makes. Rows
are identical for every system and carry the same contamination mask as §4.8.
Ours and Sójka's run at the \texttt{sojka\_val}-fitted operating points already
used there, with no re-tuning. The third-party guards run at their own default
decision, since they expose no threshold to fit, which is the same reason the
binary reduction is the only common ground available. Because the binary
reduction pools all five of our heads, the two hate-head rows of §4.8 shift
slightly here (KLEJ CBD 0.669 rather than 0.671, HateCheck-PL 0.710 rather than
0.644), while the two any-head rows reproduce exactly. Figure~\ref{fig:cross-taxonomy} plots the
result and the table beside it gives the same numbers.

\begin{longtable}[]{@{}llllll@{}}
\caption{The same comparison numerically, as \textbf{F1 / false-positive rate on negatives} per benchmark. The last column is the detection rate on all 520 \emph{Gadzi Język} prompts, which are all positive, so it is recall and not F1 and an all-positive predictor scores 1.000 on it. Generated from \texttt{reports/external\_guards/*.json} by \texttt{scripts/plot\_cross\_taxonomy\_guards.py}, so no figure here is transcribed by hand. Qwen3Guard-Gen is scored strictly, with its ``Controversial'' tier counted as safe.} \\
\toprule\noalign{}
System & KLEJ CBD & BAN-PL & PolyGuard-PL & HateCheck-PL & Gadzi detect. \\
\midrule\noalign{}
\endhead
\bottomrule\noalign{}
\endlastfoot
Baszta 1.0 (124M) & 0.669 / 0.109 & 0.667 / 0.855 & 0.562 / 0.357 & 0.710 / 0.578 & 0.906 \\
Sójka 0.1B v1.1 (124M) & 0.433 / 0.131 & 0.754 / 0.230 & 0.475 / 0.197 & 0.784 / 0.745 & 0.675 \\
HerBERT-PL-Guard (124M) & 0.459 / 0.266 & 0.762 / 0.354 & 0.816 / 0.070 & 0.786 / 0.593 & 0.992 \\
Qwen3Guard-Gen (0.6B) & 0.262 / 0.028 & 0.263 / 0.078 & 0.766 / 0.085 & 0.636 / 0.384 & 0.985 \\
\end{longtable}

\textbf{We lead on one benchmark of the four, and it is worth reporting that
plainly.} On KLEJ CBD our 0.669 is well clear of the field (HerBERT-PL-Guard
0.459, Sójka 0.433, Qwen3Guard-Gen 0.262) at the second-lowest false-positive
rate in that column, which is the same cyberbullying-without-slurs result §4.8
already identified, now measured against three more systems rather than one. On
the other three we are behind. \textbf{HerBERT-PL-Guard is the strongest system
in this table}, and it is in our own size class at 124M parameters. On
PolyGuard-PL it beats every other system on both axes at once, 0.816 F1 at a
7.0\% false-positive rate against our 0.562 at 35.7\%, and its \emph{Gadzi
Język} detection rate of 0.992 is above ours at 0.906. A reader who takes only
one number from this subsection should take that one.

\textbf{Two of those three losses are partly structural, and one is not.} The
PolyGuard-PL gap has a taxonomy-alignment component: PolyGuard's labels are
auto-mapped to Llama-Guard categories (§4.8), which is the taxonomy
HerBERT-PL-Guard was trained to emit, so it is being scored against a label
scheme built for it and against us. The BAN-PL and HateCheck-PL gaps are the
ones §4.8 already explains as a moderation-decision objective beating a
presence-of-harmful-language objective. What is not structural is our
false-positive rate. At 85.5\% on BAN-PL and 57.8\% on HateCheck-PL it is the
worst column in the table, and no framing recovers it. That is the cost of the
balanced operating point of §4.5.1, and this comparison prices it against three
systems rather than one.

\textbf{Qwen3Guard-Gen separates by task shape rather than by quality.} It is at
or near the ceiling on harmful \emph{requests} (PolyGuard-PL 0.766, \emph{Gadzi
Język} 0.985) and close to useless on Polish hate speech and cyberbullying (KLEJ
CBD 0.262, BAN-PL 0.263). The mechanism is its third safety tier. Scored
strictly, as in the table, ``Controversial'' is not a flag, and it assigns that
tier to 454 of 851 KLEJ CBD rows and 10,548 of 23,539 BAN-PL rows. Counting it
as a flag lifts BAN-PL from 0.263 to 0.623 and drives the false-positive rate
from 7.8\% to 46.8\%, so the strict reading is the one that flatters it. Neither
reading makes it a Polish hate-speech detector.

\textbf{This does not contradict the reference paper's Table 5, it measures a
different quantity.} That table reports 11.4\% precision for Qwen3Guard-Gen-0.6B
and 31.6\% for HerBERT-PL-Guard against Bielik Guard's 77.7\% on real user
traffic, where Bielik Guard's own alert rate is 2.83\%, so the prevalence of
harmful input there is a few percent. Our sets are enriched: PolyGuard-PL is
43.7\% harmful and \emph{Gadzi Język} is 100\% harmful. A broad detector with a
high alert rate scores well on F1 and recall here and falls to single-digit
precision there, and both readings are correct for the traffic they were taken
on. The ordering of systems is therefore not transferable between the two
tables, and neither table licenses a bare claim that one model is better. This
is the same lesson §4.5.1 draws from our own two operating points.

\textbf{Two limits bound this comparison.} First, taxonomy breadth is an upper
bound on the broad models' false positives rather than a measurement of them.
Llama Guard 3 and Qwen3Guard carry categories with no equivalent in the Polish
five (privacy, intellectual property, elections, specialised advice), so a row
either flags under one of those counts against it here while being correct under
its own policy. Second, and more seriously, \textbf{the contamination audit of
§4.8 covers our training pool only.} We cannot audit what HerBERT-PL-Guard,
Llama Guard 3 or Qwen3Guard-Gen were trained on, so these four sets are verified
held-out for us and for Sójka and merely presumed held-out for the rest. The
direction of that asymmetry is against us, and we report the numbers as they
came out rather than adjusting for it.

\subsection{4.9 Calibration Metrics and Reliability
Diagrams}\label{calibration-metrics-and-reliability-diagrams}

§4.3 diagnosed the OOD gap as under-confidence by \emph{eyeballing}
positive-probability means. Here we \textbf{measure} it with the
standard calibration metrics (Expected Calibration Error
\autocite{naeini2015obtaining} and its
equal-mass \emph{Adaptive}-ECE variant, Maximum Calibration Error, and
the Brier score), following the calibration treatment in Eisenstein
§4.4. Metrics are computed per category and macro-averaged over the full
520-sample \emph{Gadzi Język} set.

\begin{longtable}[]{@{}lllll@{}}
\caption{Calibration on the full 520-sample \emph{Gadzi Język} set, macro-averaged over the five categories. ECE and Adaptive-ECE (equal-mass bins) measure the mean confidence - accuracy gap, MCE the worst single bin, Brier the mean squared probability error, lower is better throughout. Temperature scaling improves Brier while slightly worsening ECE.} \\
\toprule\noalign{}
Model (Gadzi, macro over 5 cats) & ECE & Adaptive-ECE & MCE & Brier \\
\midrule\noalign{}
\endhead
\bottomrule\noalign{}
\endlastfoot
\textbf{Baszta 1.0 raw} & \textbf{0.092} & \textbf{0.096} & 0.691 & 0.075 \\
Baszta 1.0 + temperature & 0.107 & 0.111 & 0.753 & \textbf{0.068} \\
Sójka 0.1B & 0.136 & 0.139 & 0.619 & 0.114 \\
\end{longtable}

\begin{figure}
\centering
\includegraphics[width=0.95\linewidth,height=\textheight,keepaspectratio,alt={Per-category reliability diagrams for Baszta 1.0 on the Gadzi Język OOD set: mean predicted probability on the x-axis against empirical positive frequency on the y-axis, per confidence bin. Points above the diagonal indicate under-confidence, the calibration scale shift that per-category temperature scaling corrects.}]{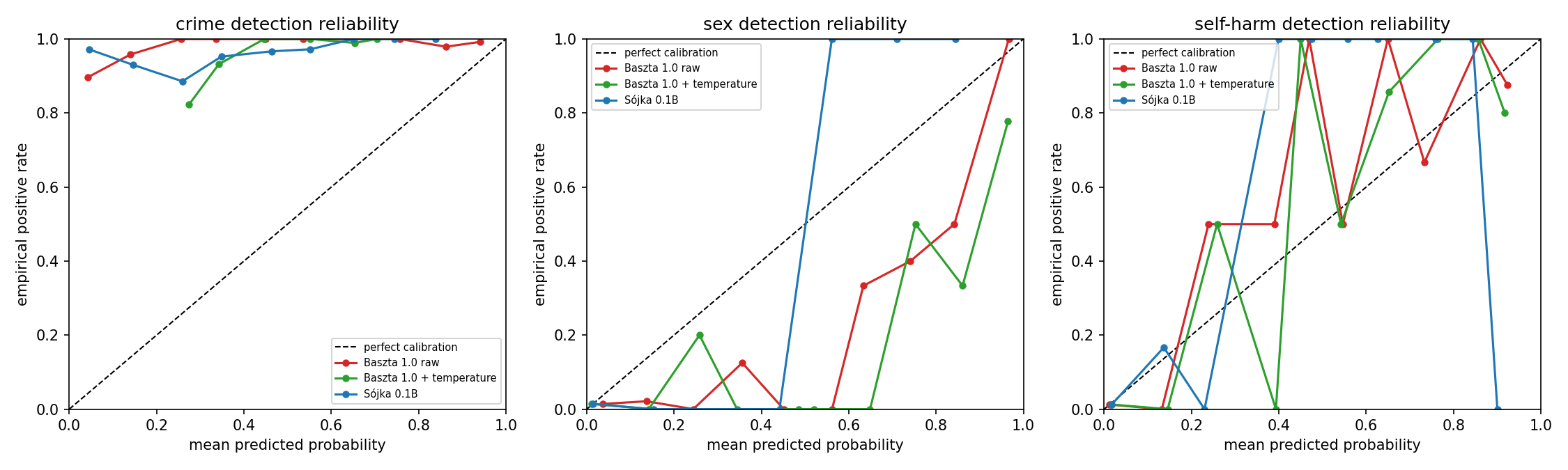}
\caption{Per-category detection reliability for \textbf{Baszta 1.0} on the
\emph{Gadzi Język} OOD set: mean predicted probability (x-axis) against
empirical positive frequency (y-axis) per confidence bin. A point
\emph{above} the diagonal is one where the observed positive rate exceeds
the predicted probability, which is the signature of under-confidence,
the calibration \emph{scale} shift that per-category temperature scaling
corrects. Points below the diagonal indicate the opposite,
over-confidence.}\label{fig:rel-baszta10}
\end{figure}

Two findings, one confirming and one qualifying earlier claims.
\textbf{(1) Our model is better calibrated on OOD than Sójka} - lower
ECE (0.092 vs.~0.136) and Brier (0.075 vs.~0.114) - which
independently supports the §4.3 reading that our OOD deficit is a
\emph{scale} problem on an otherwise well-ordered probability, not a
discrimination failure. \textbf{(2) Temperature scaling is not a free
lunch by ECE.} It lowers the Brier score (0.075 → 0.068, better
mean-squared probability) but slightly \emph{raises} ECE/MCE, because
sharpening the under-confident crime logits trades bin-level calibration
for a better decision-time operating point. This nuances §4.5's
``temperature scaling transfers'' claim: it transfers as a
\emph{threshold-enabling} rescaling (it improves F1 and Brier), not as a
global calibration improvement in the ECE sense. Reliability diagrams
for the three discriminating categories are in Figure~\ref{fig:rel-baszta10}.

\subsection{4.10 Inference-Time Deobfuscation}\label{inference-time-deobfuscation-slp-ch.-2}

The HateCheck-PL breakdown (§4.8) showed the residual failures are
surface \emph{obfuscation} - space-inserted
(\texttt{spell\_space\_add\_h} 0.41), homoglyph, and leet variants -
which defeat subword tokenization without changing meaning. Following
the normalization/tokenization treatment in Jurafsky \& Martin \autocite{jurafsky2025slp}, we add an inference-time \textbf{canonicalizer}
(Cyrillic and Greek homoglyph folding, character-spacing repair,
repeated-character collapse, and leetspeak reversal, with
URLs/emails/numbers protected) applied symmetrically before inference.
We measure it as an adversarial-robustness defence on all 520
\emph{Gadzi Język} prompts at the deployable balanced operating point: for six attacks we report
macro F1 and the \emph{flip rate} (fraction of per-label predictions
changed vs.~the clean input) without and with the defence.

\begin{longtable}[]{@{}
  >{\raggedright\arraybackslash}p{(\linewidth - 4\tabcolsep) * \real{0.3933}}
  >{\raggedright\arraybackslash}p{(\linewidth - 4\tabcolsep) * \real{0.2921}}
  >{\raggedright\arraybackslash}p{(\linewidth - 4\tabcolsep) * \real{0.3146}}@{}}
\caption{Adversarial robustness at the deployable balanced operating point (clean-input reference macro 0.496). Macro is Gadzi macro F1 under each attack. Flip is the fraction of per-label predictions that change relative to the clean input, so lower is better. Diacritics-stripping is deliberately not reversed by the canonicalizer, hence the identical columns.} \\
\toprule\noalign{}
\begin{minipage}[b]{\linewidth}\raggedright
Attack (Gadzi, balanced op-point)
\end{minipage} & \begin{minipage}[b]{\linewidth}\raggedright
No defence: macro / flip
\end{minipage} & \begin{minipage}[b]{\linewidth}\raggedright
Deobfuscated: macro / flip
\end{minipage} \\
\midrule\noalign{}
\endhead
\bottomrule\noalign{}
\endlastfoot
homoglyph & 0.477 / 0.055 & 0.496 / \textbf{0.000} \\
leetspeak & 0.415 / 0.071 & 0.471 / 0.024 \\
space-insertion & 0.409 / 0.100 & 0.485 / 0.024 \\
char-repetition & 0.296 / 0.146 & 0.476 / 0.030 \\
diacritics-strip & 0.486 / 0.026 & 0.486 / 0.026 \\
combined-heavy & 0.343 / 0.124 & 0.426 / 0.078 \\
\textbf{mean} & \textbf{0.405 / 0.087} & \textbf{0.473 / 0.030} \\
\end{longtable}

Against a clean-input macro of 0.496, the six attacks drop mean macro to
0.405 (−9 pp) and flip 8.7\% of predictions, the canonicalizer
\textbf{recovers +6.9 pp of the drop and cuts the flip rate by
two-thirds} (8.7\% → 3.0\%), fully neutralising the homoglyph attack
(flip 0) and recovering most of the character-repetition and
space-insertion damage. Diacritics-stripping is intentionally \emph{not}
reversed (it is lossy and the model is already robust to it), so it is
unchanged. This is a cheap, training-free defence that directly closes the
obfuscation failure surface §4.8 identified.

One caveat bounds the reading. Three of the six attacks (homoglyph,
leetspeak and diacritics-strip) are the same transforms used as training
augmentation in §3.1, and the canonicalizer reverses transforms we
applied ourselves. The measurement is therefore a partly closed loop,
and it establishes that the defence works against the obfuscation
families we can generate rather than against an adaptive attacker.

\subsection{4.11 Classical Bag-of-Words
Baselines}\label{classical-bag-of-words-baselines}

Both textbooks insist on a simple linear baseline as a floor
\autocite{eisenstein2019nlp}. We add
TF-IDF (word 1--2 gram + char 3--5 gram) with one-vs-rest logistic
regression and a Complement-Naive-Bayes variant, thresholds tuned on
validation only.

\begin{longtable}[]{@{}lll@{}}
\caption{Classical lexical floor: TF-IDF (word 1--2 gram + char 3--5 gram) with one-vs-rest logistic regression and Complement Naive Bayes, thresholds tuned on validation only. The HerBERT row quotes its deployable balanced operating point, whose thresholds come from \texttt{sojka\_val} rather than from the full validation split, so the three rows share a non-oracle protocol but are not fitted on identical data.} \\
\toprule\noalign{}
Baseline & Val macro / micro & Gadzi macro / micro \\
\midrule\noalign{}
\endhead
\bottomrule\noalign{}
\endlastfoot
TF-IDF + LogReg & 0.868 / 0.883 & 0.462 / 0.642 \\
TF-IDF + Compl-NB & 0.717 / 0.763 & 0.393 / 0.690 \\
\emph{HerBERT (Baszta 1.0)} & \emph{0.9111 / 0.9585 (val)} & \emph{0.496 / 0.778 (balanced
op-point)} \\
\end{longtable}

The baseline is unexpectedly informative and tempers the neural
narrative. \textbf{In distribution the task is largely lexical:} a
bag-of-words logistic regression reaches 0.868 val macro, only
\textasciitilde4.3 pp below the fine-tuned HerBERT (0.9111). \textbf{The
transformer's real value is OOD generalisation, but its edge there is
modest at a deployable operating point:} on \emph{Gadzi Język} the LogReg
baseline scores 0.462 macro versus HerBERT's 0.496 balanced-point macro
- a 3.4 pp gap - and both trail Sójka's 0.619. The transformer buys
robustness to paraphrase and domain shift that lexical features miss,
but the floor shows most of the \emph{in-distribution}
performance, and a surprising fraction of the OOD performance, is
attainable with a classical model that trains in seconds on CPU.

\subsection{4.12 Book-Grounded Training Experiments: Cost-Sensitive
Loss and
Pooling}\label{book-grounded-training-experiments-cost-sensitive-loss-and-pooling}

The above additions are evaluation-time, here we run two book-grounded
\emph{training} experiments, each a controlled variant of the Baszta 1.0 recipe
(identical data, schedule, and hyperparameters, changing only the one
factor under test), and evaluate them with the 172/348 calibration
protocol of §4.5.

\textbf{Per-class focal alpha (Baszta 1.0-α, Eisenstein §4.4.1, cost-sensitive
learning).} The Focal loss supports a per-class weight $\alpha_c$, which we set to
the inverse training frequency of each category ($\alpha_c = 1 - p_c$,
where $p_c$ is the positive rate: self-harm 0.94,
hate 0.88, crime 0.84, vulgar 0.85, sex 0.82), upweighting the positives
of rarer categories, and retrain (Baszta 1.0-α).

\begin{longtable}[]{@{}
  >{\raggedright\arraybackslash}p{(\linewidth - 10\tabcolsep) * \real{0.0959}}
  >{\raggedright\arraybackslash}p{(\linewidth - 10\tabcolsep) * \real{0.1507}}
  >{\raggedright\arraybackslash}p{(\linewidth - 10\tabcolsep) * \real{0.3014}}
  >{\raggedright\arraybackslash}p{(\linewidth - 10\tabcolsep) * \real{0.1781}}
  >{\raggedright\arraybackslash}p{(\linewidth - 10\tabcolsep) * \real{0.1233}}
  >{\raggedright\arraybackslash}p{(\linewidth - 10\tabcolsep) * \real{0.1507}}@{}}
\caption{Controlled effect of per-class focal alpha set to inverse training frequency, with data, schedule, and all other hyperparameters held fixed. OOD figures use the 172/348 protocol, lower ECE and Brier are better. Cost-sensitive weighting raises in-distribution macro while lowering OOD macro and degrading calibration.} \\
\toprule\noalign{}
\begin{minipage}[b]{\linewidth}\raggedright
Model
\end{minipage} & \begin{minipage}[b]{\linewidth}\raggedright
Val macro
\end{minipage} & \begin{minipage}[b]{\linewidth}\raggedright
Gadzi macro
\end{minipage} & \begin{minipage}[b]{\linewidth}\raggedright
Gadzi micro
\end{minipage} & \begin{minipage}[b]{\linewidth}\raggedright
OOD ECE
\end{minipage} & \begin{minipage}[b]{\linewidth}\raggedright
OOD Brier
\end{minipage} \\
\midrule\noalign{}
\endhead
\bottomrule\noalign{}
\endlastfoot
Baszta 1.0 (no alpha) & 0.9111 & \textbf{0.699} & 0.927 & \textbf{0.092} &
\textbf{0.075} \\
Baszta 1.0-α (inverse-freq alpha) & \textbf{0.9136} & 0.656 & 0.927 & 0.133 &
0.093 \\
\end{longtable}

The result is a clean instance of the paper's central tension.
Cost-sensitive weighting \textbf{improves in-distribution} validation
macro (+0.25 pp, 0.9111 → 0.9136) and lifts exactly the two categories
it most upweights - OOD crime (0.965 → 0.984) and self-harm (0.820 →
0.824) \textbf{but degrades overall OOD macro} (0.699 → 0.656,
driven by hate 0.575 → 0.439 and sex 0.606 → 0.462) and \textbf{worsens
calibration} (ECE 0.092 → 0.133, Brier 0.075 → 0.093). Upweighting rare
positives sharpens the decision boundary in a way that helps in-domain
balance and the highest-weight categories, but the extra confidence does
not transfer: it re-introduces the OOD over-sharpening that low-γ focal
was chosen to avoid. \textbf{Baszta 1.0 without alpha remains the better OOD
model}, and the micro-F1 lead over Sójka is unaffected (still
significant, p = 0.017). This reinforces §5's thesis that
in-distribution gains and OOD robustness are in tension for small
encoders under domain shift.

\emph{Independent-benchmark nuance.} Because the \emph{Gadzi Język}
macro is dominated by \emph{crime}, we re-ran Baszta 1.0-α through the
held-out external suite of §4.8, each model at its own
\texttt{sojka\_val}-fit deployable operating point. The external
evidence for cost-sensitive weighting is weaker than the OOD macro alone
suggests. Only \textbf{KLEJ CBD} improves, from 0.671 to \textbf{0.687}
hate-head F1 at a lower false-positive rate (0.102 to 0.090), which is
the upweighting of rare \emph{hate} positives working as intended.
Everywhere else the variant is worse or level: \textbf{HateCheck-PL}
falls from 0.644 to \textbf{0.613} (accuracy on hateful cases 0.54 to
0.49), \textbf{PolyGuard-PL} from 0.562 to \textbf{0.511}, and BAN-PL
is unchanged within noise (0.667 to 0.663). One improvement out of four
benchmarks is not a category rebalance, it is a narrow gain that does
not generalise, bought at the cost of OOD macro and calibration.

\textbf{Mean pooling vs.~{[}CLS{]} (Baszta 1.0-mp).} Jurafsky \& Martin
\autocite{jurafsky2025slp} note that
mask-aware mean pooling over the token sequence often beats the single
{[}CLS{]} vector \autocite{lin2023aggretriever}, especially for short inputs. We swap {[}CLS{]} for
mean pooling (Baszta 1.0-mp), keeping everything else
fixed. It gives the \textbf{best in-distribution} validation macro of
all three (0.9138) yet the \textbf{worst OOD} behaviour: \emph{Gadzi Język}
macro falls to 0.606 and critically the paired micro-F1 lead
over Sójka \textbf{loses significance} (0.913 vs.~0.903, diff +0.009, p
= 0.218), while OOD ECE rises to 0.109. The {[}CLS{]} token, though
``under-trained'' in the abstract, evidently encodes a representation
that transfers to the short imperative OOD prompts better than a mean
over the (mostly declarative) training text.

\begin{figure}
\centering
\includegraphics[width=0.95\linewidth,keepaspectratio,alt={Per-category detection reliability diagrams for Baszta 1.0-alpha, the per-class cost-sensitive variant, on the Gadzi Jezyk OOD set.}]{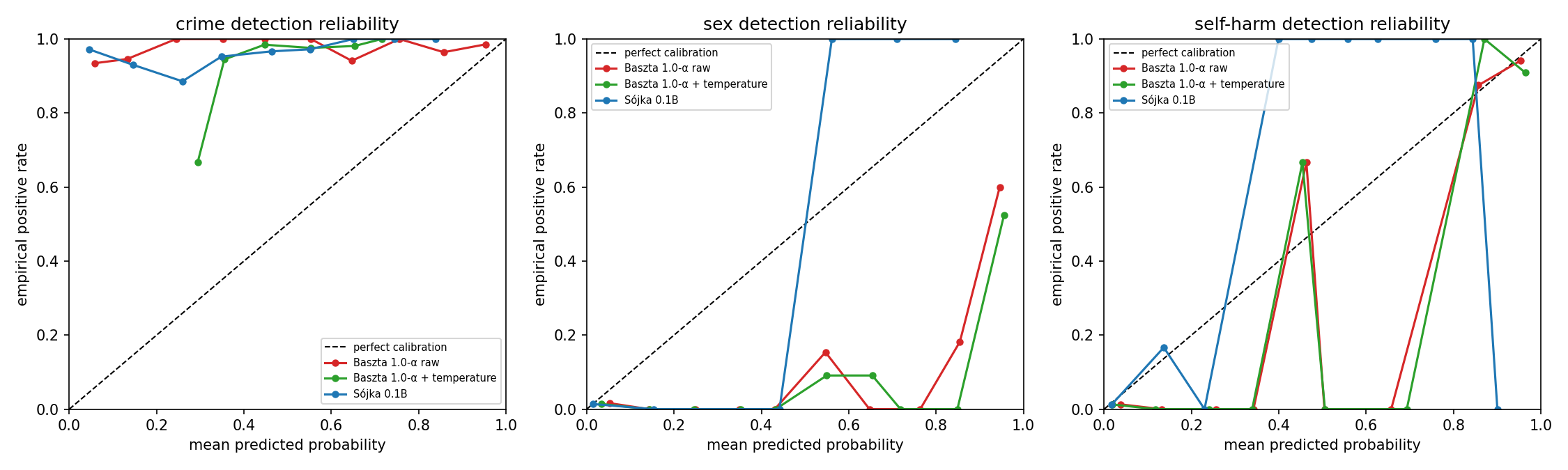}
\caption{Detection reliability for \textbf{Baszta 1.0-α} (per-class
cost-sensitive alpha). Compared with Baszta 1.0 the curves sit further
from the diagonal, which is the visual form of the ECE rise from 0.092
to 0.133 reported above.}\label{fig:rel-alpha}
\end{figure}

\begin{figure}
\centering
\includegraphics[width=0.95\linewidth,keepaspectratio,alt={Per-category detection reliability diagrams for Baszta 1.0-mp, the mean-pooling variant, on the Gadzi Jezyk OOD set.}]{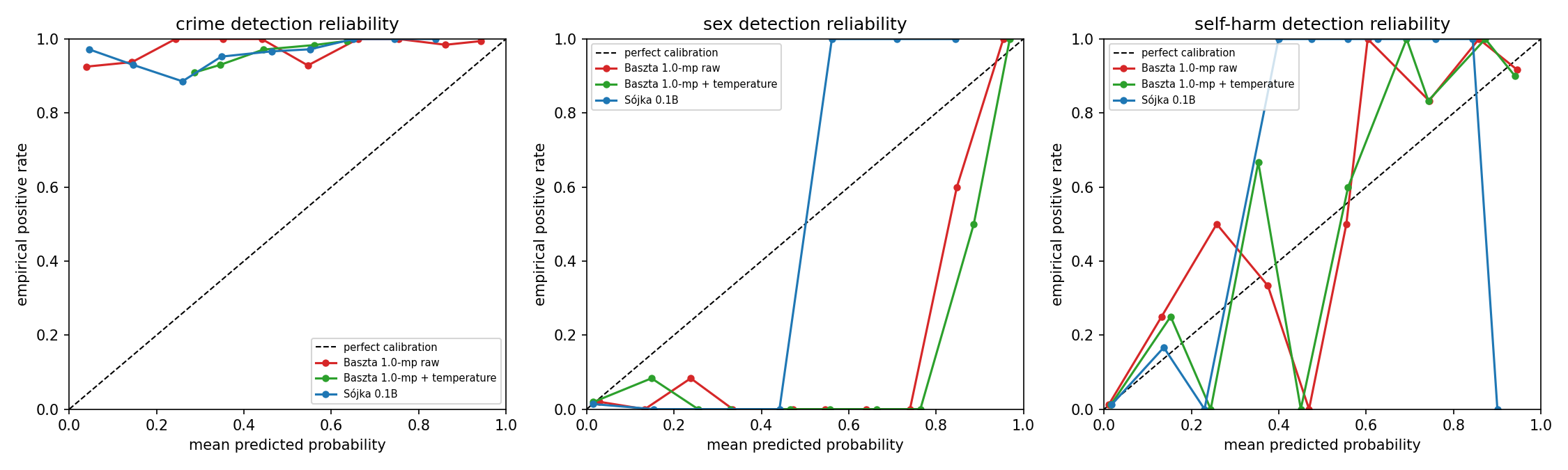}
\caption{Detection reliability for \textbf{Baszta 1.0-mp} (mean pooling),
the variant with the best validation macro and the worst OOD behaviour
of the three.}\label{fig:rel-mp}
\end{figure}

Figures~\ref{fig:rel-alpha} and~\ref{fig:rel-mp} show the reliability
curves of the two variants, for comparison with
Figure~\ref{fig:rel-baszta10}.

\textbf{Synthesis.} The three recipes rank \emph{oppositely} in- and
out-of-distribution:

\begin{longtable}[]{@{}
  >{\raggedright\arraybackslash}p{(\linewidth - 8\tabcolsep) * \real{0.1039}}
  >{\raggedright\arraybackslash}p{(\linewidth - 8\tabcolsep) * \real{0.1429}}
  >{\raggedright\arraybackslash}p{(\linewidth - 8\tabcolsep) * \real{0.2597}}
  >{\raggedright\arraybackslash}p{(\linewidth - 8\tabcolsep) * \real{0.2727}}
  >{\raggedright\arraybackslash}p{(\linewidth - 8\tabcolsep) * \real{0.2208}}@{}}
\caption{The three training recipes ranked in- versus out-of-distribution. OOD figures use the 172/348 protocol. The OOD macro column uses the 172/348 split, while the OOD micro column and its p-value come from the paired bootstrap of §4.5.2, which uses that script's own 187/333 split. This is why micro reads 0.929 here and 0.927 in the table above for the same checkpoint. Validation macro ranks the recipes in exactly the reverse order of OOD macro.} \\
\toprule\noalign{}
\begin{minipage}[b]{\linewidth}\raggedright
Recipe
\end{minipage} & \begin{minipage}[b]{\linewidth}\raggedright
Val macro
\end{minipage} & \begin{minipage}[b]{\linewidth}\raggedright
OOD macro
\end{minipage} & \begin{minipage}[b]{\linewidth}\raggedright
OOD micro vs.~Sójka
\end{minipage} & \begin{minipage}[b]{\linewidth}\raggedright
OOD ECE / Brier
\end{minipage} \\
\midrule\noalign{}
\endhead
\bottomrule\noalign{}
\endlastfoot
\textbf{Baszta 1.0} ({[}CLS{]}, no alpha) & 0.9111 & \textbf{0.699} &
\textbf{0.929 (p = 0.011 ✓)} & \textbf{0.092 / 0.075} \\
Baszta 1.0-α (per-class alpha) & 0.9136 & 0.656 & 0.927 (p = 0.017 ✓) & 0.133 /
0.093 \\
Baszta 1.0-mp (mean pooling) & \textbf{0.9138} & 0.606 & 0.913 (p = 0.218 ✗) &
0.109 / 0.093 \\
\end{longtable}

Both book-motivated changes \textbf{raise validation macro} (Baszta 1.0-mp
highest, Baszta 1.0-α second, Baszta 1.0 lowest) yet \textbf{monotonically lower OOD
macro, degrade calibration, and erode the Sójka micro-F1 lead} (Baszta 1.0-mp
loses significance entirely). This is the sharpest demonstration in the
paper of the in-distribution ↔ OOD tension: standard ``improvements''
that help the validation number actively hurt the robustness that is the
whole point of a guardrail. It also retroactively justifies the Baszta 1.0
design choices ({[}CLS{]} pooling, no per-class alpha, low-γ focal) as
an operating point selected for OOD, not for the leaderboard and
warns that tuning on in-distribution macro would have selected the
\emph{worst} OOD model.

\subsection{4.13 Deep Ensemble: Recovering Value from the Weaker
Variants}\label{deep-ensemble-recovering-value-from-the-weaker-variants}

Although Baszta 1.0-α and Baszta 1.0-mp are individually worse OOD, deep ensembling
(averaging the per-category temperature-scaled probabilities of
independently-trained models\autocite{lakshminarayanan2017simple}) can still
extract value from them (same split and paired test as §4.5.2).

\begin{longtable}[]{@{}
  >{\raggedright\arraybackslash}p{(\linewidth - 6\tabcolsep) * \real{0.2857}}
  >{\raggedright\arraybackslash}p{(\linewidth - 6\tabcolsep) * \real{0.1746}}
  >{\raggedright\arraybackslash}p{(\linewidth - 6\tabcolsep) * \real{0.3651}}
  >{\raggedright\arraybackslash}p{(\linewidth - 6\tabcolsep) * \real{0.1746}}@{}}
\caption{Deep ensembles formed by averaging per-category temperature-scaled probabilities, on the same split as §4.5.2. p-values are paired-bootstrap micro-F1 comparisons against Sójka. Lower Brier is better. The ensemble improves micro-F1 significance and probability quality but not macro, which the weaker members drag down.} \\
\toprule\noalign{}
\begin{minipage}[b]{\linewidth}\raggedright
Model / ensemble
\end{minipage} & \begin{minipage}[b]{\linewidth}\raggedright
OOD macro
\end{minipage} & \begin{minipage}[b]{\linewidth}\raggedright
OOD micro (vs.~Sójka)
\end{minipage} & \begin{minipage}[b]{\linewidth}\raggedright
OOD Brier
\end{minipage} \\
\midrule\noalign{}
\endhead
\bottomrule\noalign{}
\endlastfoot
Baszta 1.0 (single, best macro) & \textbf{0.712} & 0.929 (p = 0.011) & 0.075 \\
Baszta 1.0 + Baszta 1.0-α + Baszta 1.0-mp (mean-prob) & 0.632 & 0.932 (p = 0.005) & 0.071 \\
\textbf{Baszta 1.0 + Baszta 1.0-α (mean-prob)} & 0.660 & \textbf{0.933 (p = 0.004)} &
\textbf{0.070} \\
\end{longtable}

The ensemble delivers a \textbf{genuine, if narrow, improvement on the
two metrics that are actually defensible}: the Baszta 1.0 + Baszta 1.0-α pair raises the
\emph{significant} micro-F1 lead over Sójka from 0.929 (p = 0.011) to
\textbf{0.933 (p = 0.004)} - the strongest significance in the paper
 and gives the \textbf{best OOD Brier score (0.070)}, i.e. the most
accurate OOD probabilities of any configuration, while also lifting
self-harm F1 to 0.848. It does \textbf{not} improve macro F1 (dragged
down by the weaker members), so single-model Baszta 1.0 remains the
macro-optimal choice. The reading matches §4.5.2: macro on this
benchmark is too noisy to move reliably, but the ensemble is a real
upgrade on the metrics that carry statistical weight - micro-F1
significance and probabilistic calibration at the cost of running
three (or two) forward passes.

\textbf{Does the ensemble gain generalise?} It does not. We ran the
Baszta 1.0 + Baszta 1.0-α ensemble, with its operating point auto-fit on
\texttt{sojka\_val}, through the same held-out external suite of §4.8.
On the balanced, independent benchmarks the ensemble regresses to
Baszta 1.0 and improves nothing:

\begin{longtable}[]{@{}llll@{}}
\caption{The same three configurations on the contamination-audited external benchmarks, each at its own \texttt{sojka\_val}-fit deployable operating point with no re-tuning. The ensemble's \emph{Gadzi Język} gain does not transfer: it matches single-model Baszta 1.0 on three of four sets and is worse on PolyGuard-PL.} \\
\toprule\noalign{}
Clean held-out benchmark & Baszta 1.0 & Baszta 1.0-α & Baszta 1.0+Baszta 1.0-α ensemble \\
\midrule\noalign{}
\endhead
\bottomrule\noalign{}
\endlastfoot
HateCheck-PL hate F1 & 0.644 & 0.613 & \textbf{0.645} \\
KLEJ CBD hate F1 & 0.671 & \textbf{0.687} & 0.671 \\
BAN-PL any-head F1 & \textbf{0.667} & 0.663 & \textbf{0.667} \\
PolyGuard-PL binary F1 & \textbf{0.562} & 0.511 & 0.517 \\
\end{longtable}

Two conclusions close the arc of §4.12--§4.13. \textbf{First, the
ensemble's gain is adversarial-set-specific:} it improves micro-F1
significance and Brier on the crime-heavy, all-positive \emph{Gadzi Język}
set, but on the balanced external benchmarks its \texttt{sojka\_val}-fit
operating point collapses onto Baszta 1.0's, so it matches Baszta 1.0 exactly on
HateCheck-PL, KLEJ CBD and BAN-PL, and is \emph{slightly worse} than
Baszta 1.0 on PolyGuard-PL.
\textbf{Second, and this is the overall verdict, none of
the three new configurations (Baszta 1.0-α per-class alpha, Baszta 1.0-mp mean pooling, or
the ensemble) robustly beats Baszta 1.0 on independent external data.} Baszta 1.0-α improves one external benchmark of four and costs OOD macro and calibration. Baszta 1.0-mp is worse throughout. \textbf{The
single-model Baszta 1.0 remains the best all-around, most robust deployable
model}, and the genuine, generalising improvement of this study is the
training-free inference-time deobfuscation of §4.10, not any of the
model changes.

\section{5. Discussion}\label{discussion}

\textbf{In-distribution and OOD objectives are in tension.} The two
controlled training experiments of §4.12 make this concrete and, for a
guardrail, alarming: per-class cost-sensitive weighting (Baszta 1.0-α) and mean
pooling (Baszta 1.0-mp) both \emph{raise} validation macro above the Baszta 1.0 baseline
yet \emph{lower} OOD macro, worsen calibration, and for mean pooling
erase the significant micro-F1 lead over Sójka. Selecting a model on
the in-distribution number would have picked the worst OOD system.
Robustness must be selected for directly. It is not a by-product of
in-distribution accuracy.

\textbf{Data distribution dominates hyperparameters.} Across Baszta 0.1--0.3 (a
full sweep of LR, freeze-depth, and γ) OOD macro F1 spans only 0.4 pp,
whereas a single data-quality fix (a RefusEU multi-label loading bug)
was worth +4.6 pp, and style-matched synthetic crime data was worth
\textasciitilde40 pp on its target category. For small encoders under
domain shift, \textbf{what} the model sees matters far more than
\textbf{how} it is optimized.

\textbf{Focal loss is a double-edged sword.} It improves in-distribution
macro F1 on imbalanced data but worsens OOD calibration, an interaction
also reported by Mukhoti et al.~\autocite{mukhoti2020focal}. Lower γ (≈1.5)
trades a little in-distribution sharpness for better OOD probability
scale, a preference the completed sweep independently confirmed (best
trial γ = 1.57) and the winning Baszta 1.0 model exploited.
Pairing low-γ training with per-category temperature scaling (Section
4.5) recovers the OOD probability scale \textbf{but only when the
calibration set contains negatives.} As §4.5.1 shows, fitting the same
technique on the all-positive \emph{Gadzi Język} split produces a degenerate
operating point (crime threshold 0.025, temperature 4.3) that flags
100\% of safe text. The technique's benefit is real but is
\textbf{conditional on representative calibration data}, not automatic.
This Focal-induced OOD under-confidence, its correction by low-γ plus
temperature scaling, \emph{and} the failure mode when calibration data
lacks negatives is, we believe, the most transferable finding here,
and is not specific to Polish or to guardrails.

\textbf{Augmentation helps OOD specifically.} Holding hyperparameters
fixed, augmentation (Baszta 0.2 against its no-augmentation ablation) more than doubled the count of
OOD crime samples scored above threshold, indicating that input-space
perturbation partially simulates domain shift.

\textbf{A guardrail's headline number is a statement about its traffic.} §4.8.1
runs four other systems through the same binary decision on the same public rows
and lands in a different order from the reference paper's own cross-model table.
Both orderings are correct. Theirs is precision on a production stream that is
roughly 97\% benign, ours is F1 and false-positive rate on sets that are 44\%
and 100\% harmful, and a broad detector that looks strong on the second looks
poor on the first. The practical reading is that a guardrail cannot be selected
on a published number alone. It has to be re-measured at the prevalence it will
actually see, which is the same point §4.5.1 makes from inside our own system
about two operating points of one model.

\textbf{Generative guardrails resist self-harm generation.} The Azure
content filter blocked the majority of self-harm synthetic requests, an
ethical safety feature that nonetheless complicates \emph{defensive}
dataset construction. Template-based generation is a viable,
controllable fallback.

\section{6. Conclusion and Future
Work}\label{conclusion-and-future-work}

A data-centric HerBERT classifier holds a statistically significant but
narrow lead over Sójka on OOD micro F1, on a benchmark where micro F1 is
a weak discriminator. Under a both-tuned \textbf{paired bootstrap
test} (§4.5.2), this micro lead survives (0.929 vs.~0.903,
diff +0.026, 95\% CI {[}+0.004, +0.049{]}, p = 0.011) while the macro
lead does \textbf{not}: when Sójka is granted the same per-category
threshold-tuning our model receives, its five-way macro rises to 0.782
and our model is better in only 12.5\% of resamples. The widely-quoted
0.699-vs-0.619 macro gap was thus an \emph{operating-point artifact}
(our tuned model vs.~Sójka at a default 0.5 threshold), not a model
advantage. Nor is micro F1 a strong result on its own terms: the
always-crime baseline of §4.5.2 reaches 0.910 on the same split, which
places Sójka's tuned 0.903 below it and our 0.929 only 1.9 pp above it.
What the paper can defend is macro parity under matched tuning, a
per-category profile that is stronger on \emph{crime} and
\emph{self-harm} and weaker on \emph{sex} and \emph{vulgar}, and the
calibration results below. The decisive
modelling ingredients were \textbf{style-matched synthetic data} plus
\textbf{low-γ focal training with per-category temperature scaling},
which together convert the OOD calibration shift diagnosed \emph{and
now measured} (§4.9: our ECE 0.092 and Brier 0.075 both beat Sójka's
0.136/0.114) into a measurable gain. We stress that this gain is
\textbf{operating-point conditional}: the Gadzi benchmark has no safe
examples, so its F1-optimal point flags 100\% of safe text (§4.5.1), a
deployable point re-fit on balanced data reaches macro 0.952 at a 5.5\%
safe-text false-positive rate but gives back adversarial macro. The
completed four-phase sweep confirms that \textbf{data distribution and
lightweight calibration dominate hyperparameter choice} for small
encoders under domain shift: across 62 trials, validation macro F1
varied by under 2 pp.~Two book-grounded additions round out the picture:
a \textbf{classical TF-IDF + logistic-regression baseline} (§4.11)
reaches 0.868 val macro (within 4.3 pp of HerBERT) and 0.462 Gadzi macro
(within 3.4 pp of the deployable neural point), showing most of the
in-domain task and much of the OOD task is lexical, and a training-free
\textbf{inference-time deobfuscation} pass (§4.10) recovers
+6.9 pp of adversarial macro and cuts prediction flips by two-thirds
against surface obfuscation.

\textbf{Independent external evaluation and a data-hygiene lesson.} A
held-out, contamination-audited evaluation on four public Polish
benchmarks (§4.8) puts both systems on sets that neither of them
selected, and there the result is a two-two split rather than a lead. We
are ahead on KLEJ CBD, where the \emph{hate} head reaches 0.671
zero-shot against Sójka's 0.438, and on PolyGuard-PL. Sójka is ahead on
BAN-PL by a clear margin on both F1 and false-positive rate, because
that benchmark asks for a moderation decision on Polish social media and
is close to the distribution it was trained on, while our objective is
presence of harmful language. Its nominal lead on HateCheck-PL comes
with a 66.1\% false-positive rate on the non-hateful cases against our
33.4\%, which is the opposite of what a functional test suite is for.
The honest summary is that the two systems transfer to different tasks,
and that HateCheck-PL confirms our multi-head design routes profanity to
\emph{vulgar} rather than \emph{hate}. The audit that made these numbers trustworthy also
delivered the sharpest operational lesson here: the
otherwise-ideal PL-Guard benchmark had silently entered the training
pool (899/900) and would have produced a memorised ``0.997 detection
F1'' had we not text-matched every benchmark against the corpus
first, while the same check confirmed the headline \emph{Gadzi Język}
set is clean (0/520).

\textbf{Limitations.} Six caveats bound these claims. (1) \emph{Soft
adaptation to the test distribution:} §3.3 states plainly that the
synthetic prompts were written in the imperative jailbreak style
\emph{because that is the style of Gadzi Język}. The contamination
audit of §4.8 is exact-match, so its 0/520 result rules out copied
text but not this weaker form of adaptation. Every \emph{Gadzi Język} number
in this paper should be read with that in mind, and a fuller audit
would add approximate matching (MinHash, n-gram overlap, embedding
similarity) rather than exact matching alone. (2)
\emph{Statistical power:} the five-way macro is estimated on 520 OOD
points with one class at n = 4 and one near-universal \emph{crime}
class, so its CI is wide, the both-tuned paired test (§4.5.2) shows the macro
lead over Sójka is \textbf{not significant in either direction} (only
the micro-F1 gap is significant, p = 0.011). (3) \emph{Model-selection
bias:} our best model was chosen using OOD performance, so the
calibration/test split removes \emph{threshold} leakage but not
\emph{selection} leakage. A truly unbiased estimate needs a second,
untouched OOD set. (4) \emph{Operating-point / safe-text validity:} the
Gadzi benchmark has no safe examples, so its F1-maximising operating
point flags 100\% of safe text and is not deployable (§4.5.1). The
deployable balanced operating point (macro 0.952, 5.5\% safe-text FPR)
trades away adversarial macro (0.496), and no single point optimises
both. (5) \emph{Significance testing is uneven across the comparisons:} §4.5.2
supplies a paired bootstrap test against Sójka on \emph{Gadzi Język},
but the four-benchmark head-to-head of §4.8 reports point estimates
only. The largest of those sets (BAN-PL, n = 23,539) gives intervals
narrow enough that its outcome is not in doubt, and the KLEJ CBD gap is
wide, but the PolyGuard-PL and HateCheck-PL margins are not backed by a
paired test and should be read accordingly. (6) \emph{External-benchmark
scope and hygiene:} the held-out public results (§4.8) are moderate and
each carries a caveat. KLEJ CBD and BAN-PL are only partially disjoint
from training, with the overlapping rows dropped. BAN-PL scores a
moderation decision rather than presence of harmful language, which is a
different task from the one we trained for and which Sójka is better
suited to. PolyGuard-PL is heavily machine-translated. The one balanced,
safe-containing external set (PL-Guard) was unusable because of the
training leak. A purpose-built, uncontaminated, balanced Polish
benchmark remains the cleanest missing evaluation.

Two category-level results are worth separating from the aggregate.
Style-matched synthetic data lifts \emph{self-harm} to parity with
Sójka (0.820 against 0.815), while \emph{sex} remains the one category
where we are clearly behind under matched tuning (0.476 against 0.727),
and closing it is the most concrete modelling task this study leaves
open. Beyond that, and beyond distillation for low-latency deployment,
the most valuable experiments we did not run are:
\begin{enumerate}
    \item a threshold/temperature \emph{transfer matrix} across domains to test how
    far one calibration generalises,
    \item a \textbf{data-scaling curve}
    between the 6.9k baseline and the \textasciitilde26k augmented corpus to
    locate where synthetic data stops helping
    \item  a \textbf{synthetic-quality ablation} (random vs.~style-matched prompts)
    to isolate \emph{style matching} from sheer volume,
    \item a controlled
    head-to-head against Sójka on a freshly collected OOD set to eliminate
    the model-selection bias above. The held-out public benchmarks of §4.8
    are a first step, they validate the operating point on data
    the model was neither trained nor selected on, but they do not include
    Sójka's own predictions, so a \emph{paired}, uncontaminated,
    safe-containing Polish benchmark remains the decisive experiment.
\end{enumerate}

\section{Appendix A. Comparison Summary
vs.~Sójka}\label{appendix-a.-comparison-summary-vs.-suxf3jka}

Best model: Baszta 1.0 (low-γ focal + v2
synthetic data + per-category temperature scaling). The shared OOD
\emph{Gadzi Język} set is the only like-for-like comparison, the val row is an
internal sanity check on a \textbf{different test set} than Sójka's and
is not directly comparable. OOD figures use the 172/348
calibration split with 95\% bootstrap CIs.

\begin{longtable}[]{@{}
  >{\raggedright\arraybackslash}p{(\linewidth - 6\tabcolsep) * \real{0.2667}}
  >{\raggedright\arraybackslash}p{(\linewidth - 6\tabcolsep) * \real{0.2889}}
  >{\raggedright\arraybackslash}p{(\linewidth - 6\tabcolsep) * \real{0.2333}}
  >{\raggedright\arraybackslash}p{(\linewidth - 6\tabcolsep) * \real{0.2111}}@{}}
\caption{Head-to-head summary against Sójka 0.1B. Every \textbf{both-tuned} row gives both systems per-category tuning on the same 187-sample calibration split and scores them on the disjoint 333 samples, and those are the only rows from which a head-to-head margin should be read. The rows above them use the 172/348 split of §4.5 and leave Sójka at its default 0.5 threshold, so their large apparent margins are operating-point effects rather than model differences. Under matched tuning the per-category picture changes substantially: Sójka's \emph{crime} F1 rises from 0.587 to 0.946 and its \emph{hate} F1 from 0.045 to 0.478, which turns the previously reported +37.8 pp and +53.0 pp per-category gaps into +3.7 pp and +1.2 pp. The oracle row fits thresholds on the full test set and is an optimistic upper bound. The deployable row is in-distribution and has no Sójka counterpart, so it is not a comparison, and neither is the validation row.} \\
\toprule\noalign{}
\begin{minipage}[b]{\linewidth}\raggedright
Dimension
\end{minipage} & \begin{minipage}[b]{\linewidth}\raggedright
Ours (Baszta 1.0+temp)
\end{minipage} & \begin{minipage}[b]{\linewidth}\raggedright
Sójka 0.1B
\end{minipage} & \begin{minipage}[b]{\linewidth}\raggedright
Δ / note
\end{minipage} \\
\midrule\noalign{}
\endhead
\bottomrule\noalign{}
\endlastfoot
Encoder & HerBERT (124M) & MMLW-RoBERTa (100M) & similar \\
Loss & Focal (γ=1.5) + R-Drop & BCE & - \\
Training data & 26,248 (+synthetic) & \textasciitilde6.9K
& larger \\
Compute & 1× T4 and 2× GH200, FP16 & A100 cluster & constrained \\
\textbf{Gadzi micro F1} & \textbf{0.927} {[}0.903, 0.949{]} &
0.582 & Sójka untuned, see below \\
Gadzi macro (cal split) & 0.699 {[}0.493, 0.797{]} & 0.619 &
directional (81\% boot.) \\
Gadzi macro, 4-cat (excl. vulgar) & 0.674 {[}0.569, 0.763{]} & - &
rare-class robust \\
Gadzi macro (oracle) & 0.707 & 0.619 & optimistic UB \\
\textbf{Gadzi micro F1 (both tuned)} & \textbf{0.929} &
0.903 & \textbf{+0.026, p=0.011 sig.} \\
Gadzi macro (both tuned) & 0.712 & \textbf{0.782} & −0.070,
p=0.875 (not sig.) \\
\textbf{Deployable (balanced) macro} & \textbf{0.952} & - &
in-distribution, 5.5\% safe-text FPR \\
↳ same point, Gadzi macro & 0.496 & 0.619 & recall traded for FPR \\
Crime F1 (both tuned) & \textbf{0.983} & 0.946 & +3.7 pp \\
Hate F1 (both tuned) & \textbf{0.490} & 0.478 & +1.2 pp \\
Self-harm F1 (both tuned) & \textbf{0.812} & 0.759 & +5.3 pp \\
Sex F1 (both tuned) & 0.476 & \textbf{0.727} & −25.1 pp \\
Vulgar F1 (both tuned) & 0.800 (n = 3 test) & \textbf{1.000} & n=3
test positives, F1 estimate statistically unreliable \\
Val macro F1 (‡ not comparable) & 0.9111 & 0.770 & measured on our
internal val split, not Sójka's test set, \textbf{not a head-to-head
comparison} \\
\end{longtable}

‡The val row is reported only as an internal training sanity check. It
is measured on our own validation split, not Sójka's test split, so the
apparent gap should \textbf{not} be read as a head-to-head result. The
per-category deltas above are matched-tuning gaps against Sójka, and are
not to be confused with the \textbf{+31.0 pp} within-project crime gain
of §4.7, which measures the effect of adding synthetic data against our
own clean-data baseline. The
\textbf{deployable (balanced)} rows use the operating point of §4.5.1
(fit on a calibration set that includes safe text). They are the figures
relevant to production, whereas the Gadzi calibration-split rows
characterise adversarial recall only.

\section{Appendix B.
Reproducibility}\label{appendix-b.-reproducibility}

This appendix summarises the experimental setup at a level of detail at
which the study can be reproduced. It describes \emph{what} was done
rather than pointing at specific scripts.

\begin{itemize}
\tightlist
\item
  \textbf{Code and artifacts.} Training, evaluation, and calibration
  code, together with the experiment schedules that produced every
  reported run, are available on request. Checkpoints and the derived
  corpus are not redistributed, because the training pool includes gated
  and sensitive sources (§3.3).
\item
  \textbf{Backbone and stack.} \texttt{allegro/herbert-base-cased}
  (124M) fine-tuned with PyTorch Lightning under FP16 mixed precision on
  Python 3.11, with pinned dependency versions.
\item
  \textbf{Seeds.} Training runs use a fixed seed, and every bootstrap
  and split reported here uses seed 42. Synthetic generation is
  fixed-seed and deduplicated. Sweep trials are seeded by Optuna's own
  sampler state.
\item
  \textbf{Best model.} Low-γ (1.5) Focal + R-Drop training, followed by
  a per-category temperature and threshold fit on a held-out calibration
  split.
\item
  \textbf{Synthetic data (v2).} Style-matched imperative Polish prompts
  for the crime, sex, and self-harm categories, LLM paraphrase for
  crime, deterministic template combination for sex and self-harm,
  deduplicated and fixed-seed for reproducibility.
\item
  \textbf{Distributed sweep.} A four-phase pipeline (Optuna TPE search →
  top-5 full trainings → ten ablations → OOD \emph{Gadzi Język}
  evaluation) run against a single shared study so multiple GPUs draw
  trials without a shared filesystem.
\item
  \textbf{OOD protocol.} A disjoint 172/348 \emph{Gadzi Język}
  calibration/test split, with per-category temperatures and thresholds
  fit on the calibration side only and 95\% bootstrap confidence
  intervals (2,000 resamples, seed 42) on the test side.
\item
  \textbf{Paired significance and calibration metrics.} Both systems'
  operating points fit on the same \emph{Gadzi Język} calibration split,
  a 2,000-resample paired bootstrap of macro/micro-F1 differences
  against Sójka's recovered per-sample predictions, and ECE /
  Adaptive-ECE / MCE / Brier with reliability diagrams.
\item
  \textbf{Deployable balanced calibration.} Temperatures and thresholds
  re-fit on a balanced source that includes safe text, additionally
  reporting the false-positive rate on safe text.
\item
  \textbf{Inference-time deobfuscation.} A symmetric canonicalizer
  applied before inference - homoglyph folding, character-spacing
  repair, repeated-character collapse, and leetspeak reversal, with
  URLs, emails, and numbers protected.
\item
  \textbf{Classical baselines.} TF-IDF (word 1--2 gram + char 3--5 gram)
  with one-vs-rest logistic regression and a Complement-Naive-Bayes
  variant, thresholds tuned on validation only.
\item
  \textbf{Book-grounded training variants.} Per-class focal alpha set to
  inverse training frequency (Baszta 1.0-α) and mean vs.~{[}CLS{]} pooling (Baszta 1.0-mp),
  each a controlled clone of the best-model recipe changing only the
  factor under test.
\item
  \textbf{Deep ensemble.} Averaged per-category temperature-scaled
  probabilities of the independently trained checkpoints, evaluated on
  the same split and paired test.
\item
  \textbf{External public benchmarks.} PL-Guard, PolyGuardPrompts
  (Polish slice), HateCheck-PL, KLEJ CBD, and BAN-PL evaluated at the
  deployable operating point with \textbf{no re-tuning}, after dropping
  every row whose whitespace-normalised text appears in the training
  pool (2,000-resample bootstrap CIs, seed 42).
\end{itemize}

\printbibliography

@article{wrobel2026bielikguard,
  author  = {Wróbel, Krzysztof and Kowalski, Jan Maria and Surma, Jerzy and
             Ciuciura, Igor and Szymański, Maciej},
  title   = {{Bielik Guard}: Efficient {Polish} Language Safety Classifiers for
             {LLM} Content Moderation},
  journal = {arXiv preprint arXiv:2602.07954},
  year    = {2026},
  eprint  = {2602.07954},
  eprinttype = {arXiv},
  eprintclass = {cs.CL},
  url     = {https://arxiv.org/abs/2602.07954},
  note    = {Version 4},
}

@misc{speakleash2026sojka2,
  author       = {{SpeakLeash}},
  title        = {\texttt{speakleash/sojka-2}: Polish Content Safety Annotation Corpus},
  year         = {2026},
  howpublished = {Hugging Face dataset},
  url          = {https://huggingface.co/datasets/speakleash/sojka-2},
}

@inproceedings{kolos-etal-2024-ban,
  author    = {Kolos, Anna and Okulska, Inez and Głąbińska, Kinga and
               Karlinska, Agnieszka and Wisnios, Emilia and Ellerik, Paweł and
               Prałat, Andrzej},
  title     = {{BAN-PL}: A {Polish} Dataset of Banned Harmful and Offensive
               Content from Wykop.pl Web Service},
  booktitle = {Proceedings of the 2024 Joint International Conference on
               Computational Linguistics, Language Resources and Evaluation
               (LREC-COLING 2024)},
  editor    = {Calzolari, Nicoletta and Kan, Min-Yen and Hoste, Veronique and
               Lenci, Alessandro and Sakti, Sakriani and Xue, Nianwen},
  publisher = {ELRA and ICCL},
  address   = {Torino, Italia},
  month     = may,
  year      = {2024},
  pages     = {2107--2118},
  url       = {https://aclanthology.org/2024.lrec-main.190/},
}

@inproceedings{kolos-etal-2025-behind,
  author    = {Kołos, Anna and Lorenc, Katarzyna and Wiśnios, Emilia and
               Karlińska, Agnieszka},
  title     = {Behind Closed Words: Creating and Investigating the {forePLay}
               Annotated Dataset for {Polish} Erotic Discourse},
  booktitle = {Proceedings of the 63rd Annual Meeting of the Association for
               Computational Linguistics (Volume 1: Long Papers)},
  editor    = {Che, Wanxiang and Nabende, Joyce and Shutova, Ekaterina and
               Pilehvar, Mohammad Taher},
  publisher = {Association for Computational Linguistics},
  address   = {Vienna, Austria},
  month     = jul,
  year      = {2025},
  pages     = {2416--2432},
  doi       = {10.18653/v1/2025.acl-long.120},
  url       = {https://aclanthology.org/2025.acl-long.120/},
}

@inproceedings{ogr:kob:19:poleval,
  author    = {Ptaszyński, Michał and Pieciukiewicz, Agata and Dybała, Paweł},
  title     = {Results of the {PolEval} 2019 Shared Task 6: First Dataset and
               Open Shared Task for Automatic Cyberbullying Detection in
               {Polish} {Twitter}},
  booktitle = {Proceedings of the {PolEval} 2019 Workshop},
  editor    = {Ogrodniczuk, Maciej and Kobyliński, Łukasz},
  publisher = {Institute of Computer Science, Polish Academy of Sciences},
  address   = {Warszawa, Poland},
  year      = {2019},
  pages     = {89--110},
  isbn      = {978-83-63159-28-3},
}

@inproceedings{depotx2022devulgarization,
  author    = {Klamra, Cezary and Wojdyga, Grzegorz and Żurowski, Sebastian and
               Rosalska, Paulina and Kozłowska, Matylda and Ogrodniczuk, Maciej},
  title     = {Devulgarization of {Polish} Texts Using Pre-trained Language
               Models},
  booktitle = {Computational Science -- {ICCS} 2022},
  editor    = {Groen, Derek and de Mulatier, Clélia and Paszyński, Maciej and
               Krzhizhanovskaya, Valeria V. and Dongarra, Jack J. and
               Sloot, Peter M. A.},
  series    = {Lecture Notes in Computer Science},
  volume    = {13351},
  publisher = {Springer},
  address   = {Cham},
  year      = {2022},
  pages     = {49--55},
  doi       = {10.1007/978-3-031-08754-7_7},
}

@article{nowakowski2021detection,
  author  = {Nowakowski, Artur and Jassem, Krzysztof},
  title   = {Detection of Criminal Texts for the {Polish} State {Border Guard}},
  journal = {arXiv preprint arXiv:2108.10580},
  year    = {2021},
  eprint  = {2108.10580},
  eprinttype = {arXiv},
  eprintclass = {cs.CL},
  url     = {https://arxiv.org/abs/2108.10580},
  note    = {Presented at the 2nd International MIS2 Workshop, KDD 2021},
}

@article{krasnodebska2026multilingual,
  author  = {Krasnodębska, Aleksandra and Kusa, Wojciech and Lipani, Aldo},
  title   = {Multilingual Refusal Alignment for Safer Large Language Models},
  journal = {arXiv preprint arXiv:2606.07535},
  year    = {2026},
  eprint  = {2606.07535},
  eprinttype = {arXiv},
  eprintclass = {cs.CL},
  url     = {https://arxiv.org/abs/2606.07535},
  note    = {Findings of the Association for Computational Linguistics: ACL 2026},
}

@inproceedings{lin2017focal,
  author    = {Lin, Tsung-Yi and Goyal, Priya and Girshick, Ross and
               He, Kaiming and Dollár, Piotr},
  title     = {Focal Loss for Dense Object Detection},
  booktitle = {Proceedings of the {IEEE} International Conference on Computer
               Vision ({ICCV})},
  year      = {2017},
  pages     = {2980--2988},
}

@inproceedings{liang2021rdrop,
  author    = {Liang, Xiaobo and Wu, Lijun and Li, Juntao and Wang, Yue and
               Meng, Qi and Qin, Tao and Chen, Wei and Zhang, Min and
               Liu, Tie-Yan},
  title     = {{R-Drop}: Regularized Dropout for Neural Networks},
  booktitle = {Advances in Neural Information Processing Systems ({NeurIPS})},
  volume    = {34},
  year      = {2021},
  pages     = {10890--10905},
}

@inproceedings{lakshminarayanan2017simple,
  author    = {Lakshminarayanan, Balaji and Pritzel, Alexander and
               Blundell, Charles},
  title     = {Simple and Scalable Predictive Uncertainty Estimation Using Deep
               Ensembles},
  booktitle = {Advances in Neural Information Processing Systems ({NIPS})},
  volume    = {30},
  year      = {2017},
  pages     = {6402--6413},
}

@article{boken2021appropriateness,
  author  = {Böken, Björn},
  title   = {On the Appropriateness of {Platt} Scaling in Classifier
             Calibration},
  journal = {Information Systems},
  volume  = {95},
  pages   = {101641},
  year    = {2021},
}

@article{zou2023universal,
  author  = {Zou, Andy and Wang, Zifan and Carlini, Nicholas and Nasr, Milad and
             Kolter, J. Zico and Fredrikson, Matt},
  title   = {Universal and Transferable Adversarial Attacks on Aligned Language
             Models},
  journal = {arXiv preprint arXiv:2307.15043},
  year    = {2023},
  eprint  = {2307.15043},
  eprinttype = {arXiv},
  eprintclass = {cs.CL},
  url     = {https://arxiv.org/abs/2307.15043},
}

@book{jurafsky2025slp,
  author    = {Jurafsky, Dan and Martin, James H.},
  title     = {Speech and Language Processing: An Introduction to Natural
               Language Processing, Computational Linguistics, and Speech
               Recognition with Language Models},
  edition   = {3},
  year      = {2025},
  note      = {Online manuscript released January 12, 2025},
  url       = {https://web.stanford.edu/~jurafsky/slp3/},
}

@book{eisenstein2019nlp,
  author    = {Eisenstein, Jacob},
  title     = {Introduction to Natural Language Processing},
  publisher = {MIT Press},
  address   = {Cambridge, MA},
  year      = {2019},
  isbn      = {978-0-262-04284-0},
}

@inproceedings{mroczkowski2021herbert,
  author    = {Mroczkowski, Robert and Rybak, Piotr and Wróblewska, Alina and
               Gawlik, Ireneusz},
  title     = {{HerBERT}: Efficiently Pretrained Transformer-based Language
               Model for {Polish}},
  booktitle = {Proceedings of the 8th Workshop on Balto-Slavic Natural Language
               Processing (BSNLP)},
  publisher = {Association for Computational Linguistics},
  year      = {2021},
  pages     = {1--10},
}

@inproceedings{rybak2020klej,
  author    = {Rybak, Piotr and Mroczkowski, Robert and Tracz, Janusz and
               Gawlik, Ireneusz},
  title     = {{KLEJ}: Comprehensive Benchmark for {Polish} Language
               Understanding},
  booktitle = {Proceedings of the 58th Annual Meeting of the Association for
               Computational Linguistics},
  publisher = {Association for Computational Linguistics},
  year      = {2020},
  pages     = {1191--1201},
}

@inproceedings{dadas2024pirb,
  author    = {Dadas, Sławomir and Perełkiewicz, Michał and Poświata, Rafał},
  title     = {{PIRB}: A Comprehensive Benchmark of {Polish} Dense and Hybrid
               Text Retrieval Methods},
  booktitle = {Proceedings of the 2024 Joint International Conference on
               Computational Linguistics, Language Resources and Evaluation
               (LREC-COLING 2024)},
  publisher = {ELRA and ICCL},
  year      = {2024},
  pages     = {12761--12774},
}

@inproceedings{mukhoti2020focal,
  author    = {Mukhoti, Jishnu and Kulharia, Viveka and Sanyal, Amartya and
               Golodetz, Stuart and Torr, Philip H. S. and Dokania, Puneet K.},
  title     = {Calibrating Deep Neural Networks Using Focal Loss},
  booktitle = {Advances in Neural Information Processing Systems ({NeurIPS})},
  volume    = {33},
  year      = {2020},
  pages     = {15288--15299},
}

@article{lin2023aggretriever,
  author  = {Lin, Sheng-Chieh and Li, Minghan and Lin, Jimmy},
  title   = {{Aggretriever}: A Simple Approach to Aggregate Textual
             Representations for Robust Dense Passage Retrieval},
  journal = {Transactions of the Association for Computational Linguistics},
  volume  = {11},
  year    = {2023},
  pages   = {436--452},
}

@inproceedings{krasnodebska2025plguard,
  author    = {Krasnodębska, Aleksandra and Seweryn, Karolina and
               Łukasik, Szymon and Kusa, Wojciech},
  title     = {{PL-Guard}: Benchmarking Language Model Safety for {Polish}},
  booktitle = {Proceedings of the 10th Workshop on Slavic Natural Language
               Processing (Slavic NLP 2025)},
  publisher = {Association for Computational Linguistics},
  address   = {Vienna, Austria},
  year      = {2025},
  pages     = {25--37},
  url       = {https://aclanthology.org/2025.bsnlp-1.4/},
}

@article{kumar2025polyguard,
  author  = {Kumar, Priyanshu and Jain, Devansh and Yerukola, Akhila and
             Jiang, Liwei and Beniwal, Himanshu and Hartvigsen, Thomas and
             Sap, Maarten},
  title   = {{PolyGuard}: A Multilingual Safety Moderation Tool for 17
             Languages},
  journal = {arXiv preprint arXiv:2504.04377},
  year    = {2025},
  eprint  = {2504.04377},
  eprinttype = {arXiv},
  eprintclass = {cs.CL},
  url     = {https://arxiv.org/abs/2504.04377},
}

@inproceedings{rottger2022multilingual,
  author    = {Röttger, Paul and Seelawi, Haitham and Nozza, Debora and
               Talat, Zeerak and Vidgen, Bertie},
  title     = {Multilingual {HateCheck}: Functional Tests for Multilingual Hate
               Speech Detection Models},
  booktitle = {Proceedings of the Sixth Workshop on Online Abuse and Harms
               (WOAH)},
  publisher = {Association for Computational Linguistics},
  year      = {2022},
  pages     = {154--169},
}

@inproceedings{guo2017calibration,
  author    = {Guo, Chuan and Pleiss, Geoff and Sun, Yu and
               Weinberger, Kilian Q.},
  title     = {On Calibration of Modern Neural Networks},
  booktitle = {Proceedings of the 34th International Conference on Machine
               Learning ({ICML})},
  year      = {2017},
  pages     = {1321--1330},
}

@incollection{platt1999probabilistic,
  author    = {Platt, John},
  title     = {Probabilistic Outputs for Support Vector Machines and
               Comparisons to Regularized Likelihood Methods},
  booktitle = {Advances in Large Margin Classifiers},
  publisher = {MIT Press},
  year      = {1999},
  pages     = {61--74},
}

@inproceedings{naeini2015obtaining,
  author    = {Naeini, Mahdi Pakdaman and Cooper, Gregory F. and
               Hauskrecht, Milos},
  title     = {Obtaining Well Calibrated Probabilities Using {Bayesian}
               Binning},
  booktitle = {Proceedings of the 29th {AAAI} Conference on Artificial
               Intelligence},
  year      = {2015},
  pages     = {2901--2907},
}

@inproceedings{bergkirkpatrick2012empirical,
  author    = {Berg-Kirkpatrick, Taylor and Burkett, David and Klein, Dan},
  title     = {An Empirical Investigation of Statistical Significance in {NLP}},
  booktitle = {Proceedings of the 2012 Joint Conference on Empirical Methods in
               Natural Language Processing and Computational Natural Language
               Learning},
  publisher = {Association for Computational Linguistics},
  year      = {2012},
  pages     = {995--1005},
}

@misc{llamateam2024llama3,
  author  = {Grattafiori, Aaron and Dubey, Abhimanyu and Jauhri, Abhinav and
             others},
  title   = {The {Llama 3} Herd of Models},
  year    = {2024},
  eprint  = {2407.21783},
  eprinttype = {arXiv},
  eprintclass = {cs.AI},
  url     = {https://arxiv.org/abs/2407.21783},
  note    = {{Llama Guard 3} is released with this model family},
}

@misc{zhao2025qwen3guard,
  author  = {Zhao, Haiquan and Yuan, Chenhan and Huang, Fei and Hu, Xiaomeng
             and Zhang, Yichang and Yang, An and Yu, Bowen and Liu, Dayiheng
             and others},
  title   = {{Qwen3Guard} Technical Report},
  year    = {2025},
  eprint  = {2510.14276},
  eprinttype = {arXiv},
  eprintclass = {cs.CL},
  url     = {https://arxiv.org/abs/2510.14276},
}

\end{document}